\documentclass[lettersize,journal]{IEEEtran}
\usepackage{amsmath,amsfonts}
\usepackage{algorithmic}
\usepackage{algorithm}
\usepackage{array}
\usepackage[caption=false,font=normalsize,labelfont=sf,textfont=sf]{subfig}
\usepackage{textcomp}
\usepackage{stfloats}
\usepackage{url}
\usepackage{verbatim}
\usepackage{graphicx}
\usepackage{cite}
\usepackage{amssymb,amsthm,bm}
\usepackage{booktabs}

\usepackage{multirow}
\usepackage{breqn}

\DeclareMathOperator*{\sinc}{sinc}
\DeclareMathOperator*{\sgn}{sgn}

\begin{document}

\title{Efficient Collision Detection Oriented Motion Primitives for Path Planning}
\author{DallaLibera~Fabio$^{1}$, Abe~Shinya$^{1}$, and Ando~Takeshi$^{1}$%
  \thanks{$^{1}$Panasonic Holdings Corporation, Manufacturing Innovation Division, Robotics Promotion Office, Kadoma City, Osaka, Japan
        {\tt\footnotesize dallalibera.fabio@jp.panasonic.com}}%
\thanks{Digital Object Identifier (DOI): see top of this page.}
}

\maketitle

\begin{abstract}
  Mobile robots in dynamic environments require fast planning, especially when onboard computational resources are limited. While classic potential field based algorithms may suffice in simple scenarios, in most cases algorithms able to escape local minima are necessary.
  Configuration-space search algorithms have proven to provide a good trade-off between quality of the solutions and search time. Literature presents a wide variety of approaches that speed up this search by reducing the number of edges that need to be inspected. Much less attention was instead given to reducing the time necessary to evaluate the cost of a single edge.
  This paper addresses this point by associating edges to motion primitives that prioritize fast collision detection.
  We show how biarcs can be used as motion primitives that enable fast collision detection, while still providing smooth, tangent continuous paths. The proposed approach does not assume a disc shaped hitbox, making it appealing for all robots with very different width and length or for differential drive robots with active wheels located far from the robot's center.
\end{abstract}


\IEEEpeerreviewmaketitle

\section{Introduction}

\IEEEPARstart{P}{ath} planning has been a subject of extensive research in the field of robotics, yielding a diverse array of approaches over the years~\cite{mac2016heuristic,vagale2021path,sun2021motion}. In cases where stringent computational efficiency is a primary concern, classic reactive strategies, exemplified by methods like the Virtual Potential Field~\cite{borenstein1989real} and the Vector Field Histogram~\cite{borenstein1991vector}, as well as their subsequent developments~\cite{ulrich1998vfh+,minguez2004nearness}, may offer viable solutions.

Nonetheless, in numerous scenarios, these methodologies are unsuitable due to their susceptibility to entrapment in local minima. Alternative potential field formulations, which alleviate this concern, have been put forth in the literature~\cite{singh1996real,connolly1990path,heidari2021collision}.  However, it is important to note that the practical utility of these alternatives is often confined to specific contexts, or their adoption is associated with a substantial escalation in computational overhead~\cite{agirrebeitia2005new}.

To address the inherent challenge of limited path information in potential field algorithms, an array of local search approaches has emerged over the years. This category of methods typically involves the iterative adjustment of a path, frequently represented as splines~\cite{connors2007analysis,lau2009kinodynamic,sprunk2011online,sprunk2012improved}. Some notable approaches within this realm include Elastic Bands~\cite{quinlan1993elastic}, Reactive Path Deformation~\cite{lamiraux2004reactive}, Trajectory Deformer~\cite{kurniawati2007from,fraichard2009navigating}, and Timed Elastic Band~\cite{rosmann2013efficient}. Although these methods are generally more robust and sufficiently efficient when compared to potential field algorithms, they tend to yield sub-optimal solutions.

Alternatively, there exist other techniques, such as the Dynamic Window Approach~\cite{fox1997dynamic} and its continuous-space extensions~\cite{brock1999high,simmons1996curvature,quasny2003curvature}, which conduct local searches within the velocity search space. These methods, particularly when combined with a global planner, often demonstrate a high degree of robustness in real life scenarios. In recent years, machine learning techniques, particularly deep learning algorithms, have demonstrated remarkable performance in complex environments without the need for exhaustive search processes, as evidenced by works such as~\cite{rabiee2019ivoa,wang2021agile,xie2023drl}. However, a notable limitation is that, in the absence of dedicated hardware acceleration, even basic inference operations can incur significant computational expenses.

An alternative strategy for improving the planning results involves the transformation of the configuration space into a convex space, conducting the planning within this convex space, and subsequently mapping the solution back to the original configuration space. It is essential to note, however, that this approach is most suitable for static environments, wherein the computationally intensive mapping definition is executed only once~\cite{suryawanshi2003domain}.

Discretizing the configuration space serves to convert the intrinsically continuous problem of path planning into a graph search problem. A range of strategies for selecting the sampling approach has been thoroughly examined, encompassing widely recognized methods such as the Rapidly Exploring Random Tree (RRT) algorithm~\cite{lavalle1998rapidly} and its derivatives~\cite{noreen2016optimal}, as well as the Probabilistic Road Map (PRM) approach~\cite{kavraki1996probabilistic}. The sampling procedures employed by these algorithms are geared towards reducing the set of configurations that must be explored before a path between the start and goal nodes is identified. In terms of the underlying graph, the primary objective is to minimize the number of nodes that constitute the graph. Aspects of the path, such as its smoothness, are often addressed through successive refinement steps~\cite{li2022smooth} or are managed through deterministic optimizations integrated with the stochastic sampling process~\cite{essaidi2022minimum}. 

Another line of investigation has focused on minimizing the search cost within a given graph. In many motion planning scenarios, effective heuristics are available for A* to significantly reduce the number of nodes traversed during a search compared to Dijkstra's algorithm. Additionally, in the majority of cases, obstacle positions are updated as new sensor data becomes available. Rather than conducting an entirely new search, prior computation results are leveraged to expedite the search process. This approach is adopted by algorithms such as D*~\cite{stentz1997optimal}, Lifelong Planning A*~\cite{koenig2001incremental}, and D* Lite~\cite{koenig2002d}.

An additional research direction has been dedicated to reducing the number of evaluated edges. Specifically, the actual cost of an edge is computed only when absolutely necessary, and heuristics are employed as substitutes for the actual computations whenever feasible. This approach is exemplified in algorithms such as Lazy PRM~\cite{bohlin2000path}, LWA*~\cite{cohen2014planning}, Lazy-PRM* and Lazy RRG*~\cite{hauser2015lazy}, LazySP~\cite{dellin2016unifying}, and Lifelong-GLS~\cite{lim2022lazy}.

We employ an alternative approach to enhance the efficiency of edge cost evaluation. In the physical realm, the edges on the graph correspond to the robot's movements, commonly referred to as motion primitives. While path smoothness and path length have traditionally been the primary design criteria for motion primitives, this paper prioritizes collision detection efficiency as the foremost design criterion. Through the precise design of motion primitives, we enable accelerated path planning without a significant compromise on path characteristics.

Section~\ref{sec:biarcDef} briefly introduces biarcs, the curves utilized to generate motion primitives in this study. Section~\ref{sec:planJ} examines the impact of a free parameter on path length and path smoothness and identifies an optimal value for achieving desirable paths. Section~\ref{sec:replanJ} concentrates on replanning and discusses a heuristic for determining the free parameter in this context. Section~\ref{sec:collisionDetection} elaborates on the efficient execution of collision detection. Section~\ref{sec:experiment} presents the experimental results. Lastly, in Section~\ref{sec:conclusions}, the paper concludes by assessing the limitations of the proposed approach and outlining potential avenues for future research.

\section{Biarcs}
\label{sec:biarcDef}

Biarcs have been the subject of extensive investigation in multiple domains, including Mathematics~\cite{sandel1937geometrie,meek2008family}, Computer Graphics~\cite{kim2010precise,wang1993orientation}, and, particularly, in the field of Computer-Aided Design~\cite{bolton1975biarc,sabin1977use,meek1992approximation,parkinson1992optimised,walton1996approximation,kurnosenko2013biarcs}. One significant motivation for the extensive exploration of biarcs in the latter domain is the common provision of circular interpolation algorithms in the control modules of early numerically controlled manufacturing systems. Biarc trajectories, as a result, hold appeal as readily usable tool paths.

Biarcs as paths in the field of robotics have received significantly less attention~\cite{naik2005arc,ren2020circular,arita2020optimal}. One contributing factor is that biarcs exhibit $G_1$ continuity, whereas $G_2$ continuity is a prerequisite for Ackermann steering. However, it is worth noting that while $G_2$ continuity is a desirable trait, it is not essential for many types of robots, including holonomic robots, differential drive robots, or skid steering robots. In fact, for these categories of robots, even $G_0$ continuity can be sufficient.
Indeed, straight paths and in-place rotations are often employed in robotics because they typically represent the shortest path~\cite{wurman2008coordinating}.  Nonetheless, this type of motion necessitates the robot to come to a complete stop at each waypoint, leading to frequent acceleration and deceleration, which can result in increased travel time and wear and tear. As a result, $G_1$ continuous curves, specifically biarcs, provide an appealing compromise between smoothness and path planning simplicity

To the best of our knowledge, there are no documented instances of the use of biarcs as motion primitives for graph-based planners. Additionally, despite the maturity of the biarc literature, it often overlooks factors crucial for robotics, such as path length and degenerate cases~\cite{nutbourne1988differential,su2014computational}. This section seeks to address this gap by presenting an analysis concentrated on aspects pertinent to path planning.

\begin{figure}%
\centering
\subfloat[][]{%
\label{fig:biarcc}%
    \includegraphics[width=0.46\linewidth]{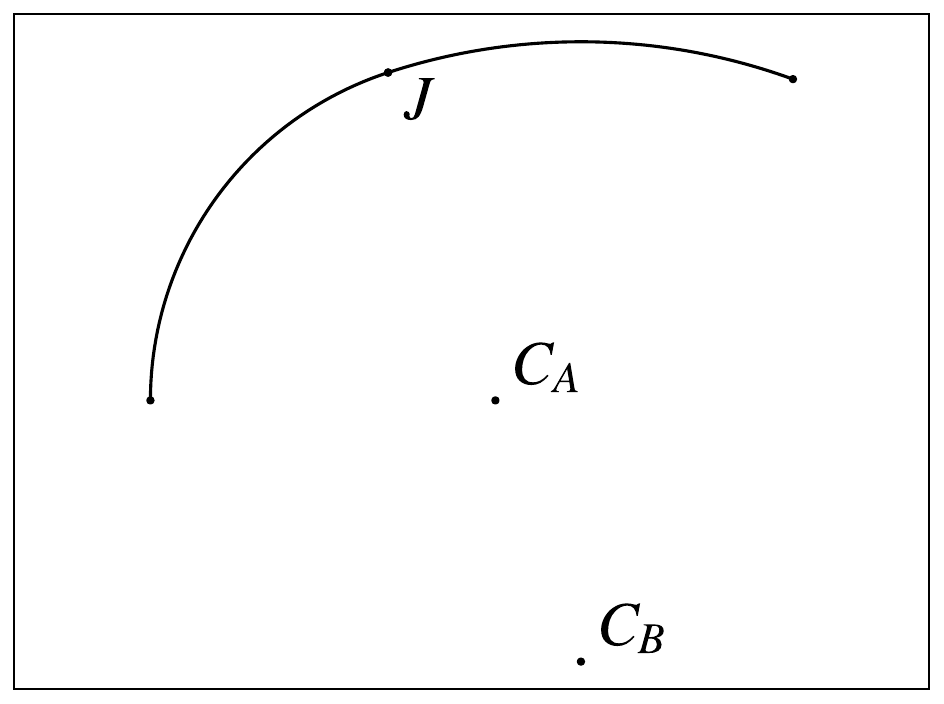}  }%
\hspace{8pt}%
\subfloat[][]{%
\label{fig:biarcs}%
    \includegraphics[width=0.46\linewidth]{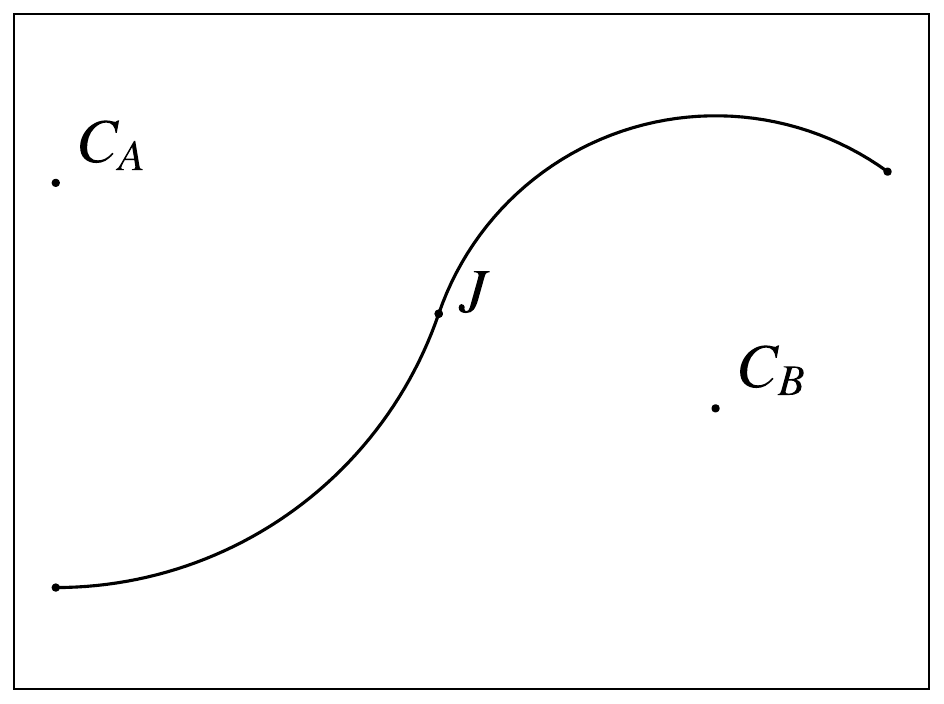}  }\\
\subfloat[][]{%
\label{fig:biarcj}%
    \includegraphics[width=0.46\linewidth]{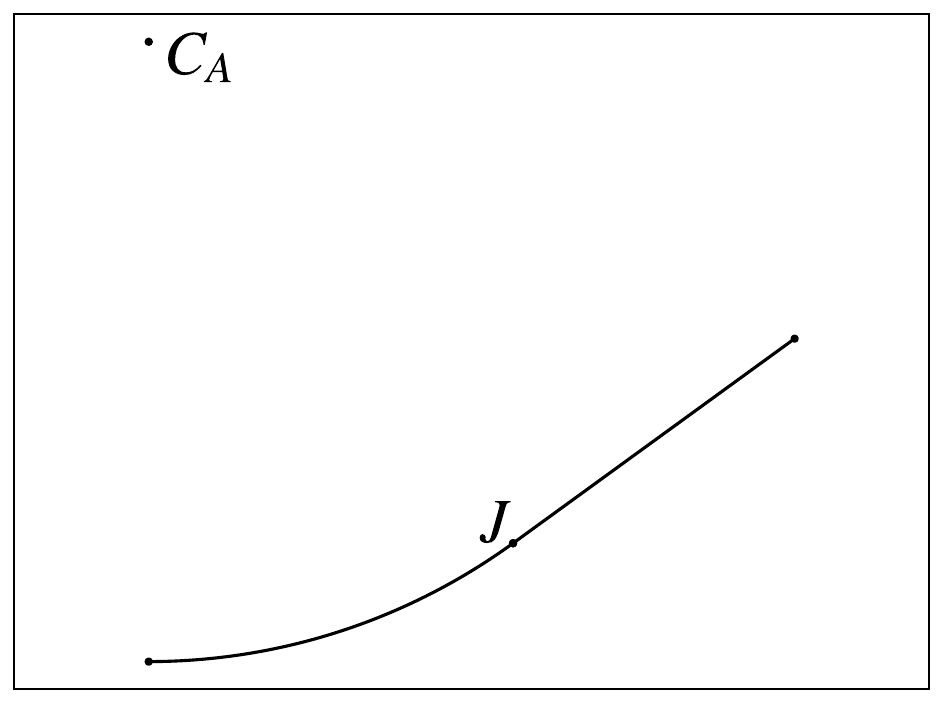}  }%
\hspace{8pt}%
\subfloat[][]{%
\label{fig:biarceq}%
    \includegraphics[width=0.46\linewidth]{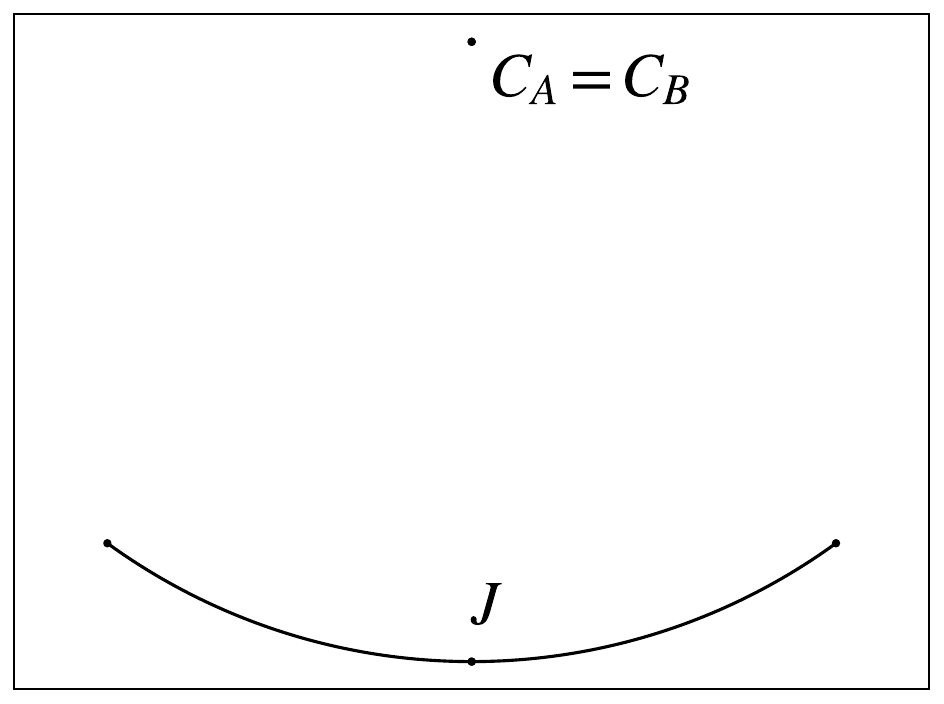}  }%
\caption{Examples of biarc curves. $C_A$ and $C_B$ indicate the two arcs centers. The joint point is denoted by $J$. In panel (c) the point $C_B$ is at infinity.}%
\label{fig:biarcExamples}%
\end{figure}

A biarc is a curve comprising two arcs of circles, as illustrated in Figure~\ref{fig:biarcs}. The centers of rotation for these two arcs are denoted as $C_A$ and $C_B$, and the point at which the two arcs intersect is marked as joint point $J$. The choice of the two arcs is made so that their tangents at point $J$ are collinear. The centers of rotation for the two arcs can be located either on the same side of the curve (as depicted in Figure~\ref{fig:biarcc}) or on opposite sides of the curve (as depicted in Figure~\ref{fig:biarcs}). A special case arises when the radius of rotation is infinite, resulting in a straight segment (as shown in Figure~\ref{fig:biarcj}). It is worth noting that both arcs can become straight segments, and, therefore, line segments are a subset of the biarc family. Lastly, when the two centers of rotation coincide (as seen in Figure~\ref{fig:biarceq}), the biarc is equivalent to a single arc of a circle.

Biarcs represent one of the simplest curve types that enable the specification of both initial position and orientation as well as final position and orientation. Notably, in the case of a single arc, it is not feasible to define the arrival orientation once the arrival position is determined. For biarcs, when provided with an initial position $A$ and orientation $\theta_A$, and a final position $B$ and orientation $\theta_B$ there exists an infinite array of options for the selection of the joint point $J$. To elaborate, let us establish the following definitions (as illustrated in Figure~\ref{fig:biarcDefs}):

\begin{figure}
\centering

  \includegraphics[width=\linewidth]{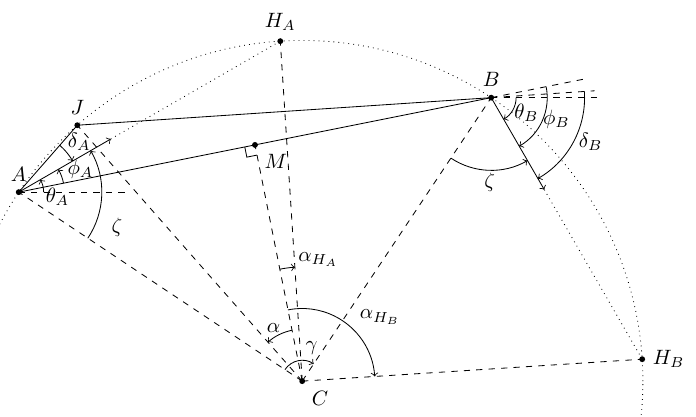}  
  \caption{ Notation used throughout the paper. $\theta_A$ and $\theta_B$ are the orientations with respect to the global frame, $\phi_A$ and $\phi_B$ are the orientations with respect to the chord connecting the initial and final position, $\delta_A$ and $\delta_B$ are the orientations with respect to the chords connecting the biarc extremes to the joint point. $\zeta$ is the orientation with respect to the radius, $\gamma$ is the difference between the final and initial orientation, $\alpha$ is used to parametrize the joint location, and $\alpha_{H_A}$ and $\alpha_{H_B}$ are the angles for which the joint is placed along the initial and final orientation, respectively.}
  \label{fig:biarcDefs}
\end{figure}

\begin {itemize}
\item $-\pi < \phi_A \leq \pi$ as the angle from the segment $AB$ to the orientation of the robot in $A$
\item $-\pi < \phi_B \leq \pi$ as the angle from the segment $AB$ to the orientation of the robot in $B$
\item $\gamma = \phi_B-\phi_A$
\item The unit vector $\bm{u}=\left[\begin{matrix} u_x\\ u_y \end{matrix} \right]$ as the direction from $A$ to $B$
\item A unit vector $\bm{v}$ orthogonal to $\bm{u}$ as  $\bm{v}=\left[\begin{matrix} -u_y\\ u_x \end{matrix} \right]$
\item The length of $AB$ as $|AB|$
\item The midpoint of $AB$ as $M$, i.e. $M=A+\frac{|AB|}{2}\bm{u}$
\item The point $C$ as
  \begin{eqnarray}
    C&=& \begin{cases} M+ \frac{|AB|}{2}\frac{\bm{v}}{\tan\left( \frac{\gamma}{2}\right)} & \text{if } \gamma \neq k \pi$, $k\in \mathbb{Z}\\
           M  & \text{if }\gamma = \pi+ k 2\pi
         \end{cases}
           \label{eq:cpos}
  \end{eqnarray}
\end{itemize}

\begin{figure}
\centering
  \includegraphics[width=0.8\linewidth]{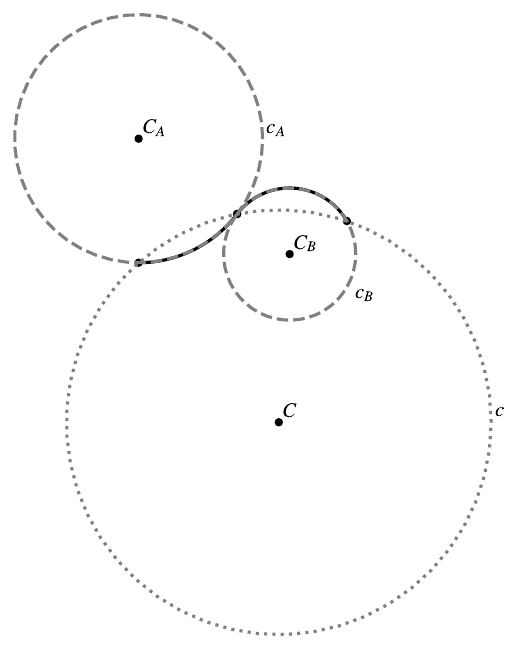}  
  \caption{ A biarc is composed of two arcs. The first arc is part of a circle $c_A$ with center $C_A$. The second arc is part of a circle $c_B$ with center $C_B$. The locus of the joint point is a circle $c$ with center $C$. In the limit case, each of these three circles can have an infinite radius and become a line.}
  \label{fig:biarcCircles}
\end{figure}

If the initial and final directions are different, i.e., if $\gamma \neq k 2\pi$, where $k \in \mathbb{Z}$, then $J$ can be selected from any of the points situated on the circle with its center at $C$ and a radius of $|AC|=|BC|$. An example of this is illustrated in Figure~\ref{fig:biarcCircles}. Detailed proofs can be found in~\cite{nutbourne1988differential}. In case the directions are the same ($\gamma = k2\pi$), then $J$ can be chosen from any point along the line $AB$.

\begin{figure}
\centering
\subfloat[][]{%
\label{fig:biarcAlpha}%
\includegraphics[width=\linewidth]{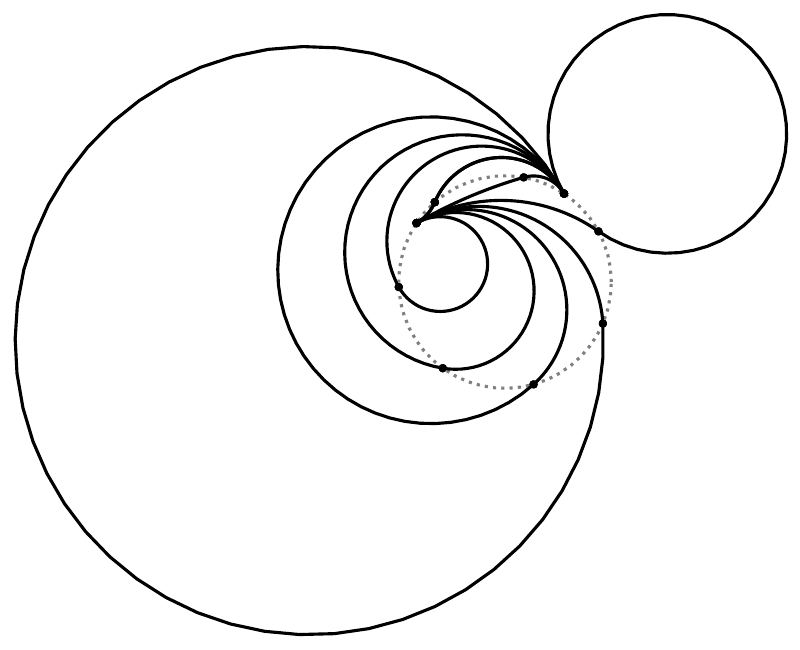}  }\\%
\hspace{8pt}%
\subfloat[][]{%
\label{fig:biarcAlphaColl}%
\includegraphics[width=\linewidth]{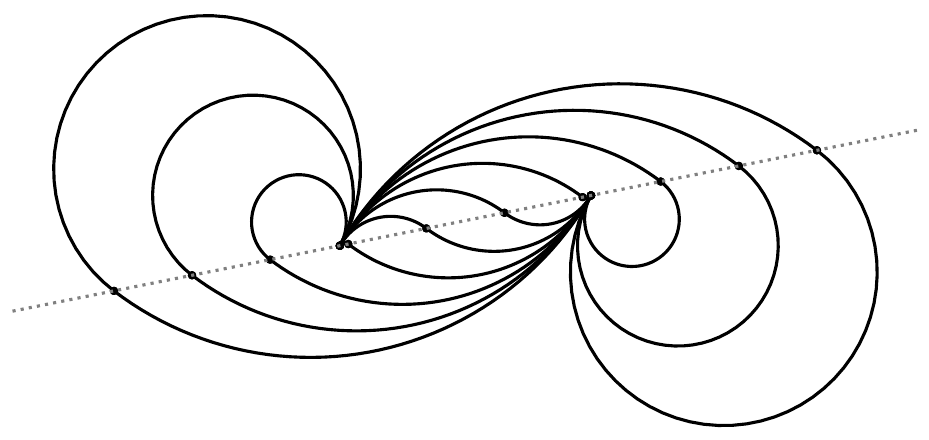}  }
\caption{
Examples of potential trajectories between a specific pair of initial and final positions and orientations. (a) illustrates the non-degenerate case ($\gamma \neq k2\pi$, $k\in \mathbb{Z}$), while (b) presents the degenerate case ($k\in \mathbb{Z}$). It is noteworthy that the degenerate case frequently occurs in practical scenarios, where the initial or final orientations are often coincident.}
  \label{fig:biarcAlphaAll}
\end{figure}

Figure~\ref{fig:biarcAlpha} illustrates how various biarcs, corresponding to different choices of $J$, can be employed to transition from an initial position and orientation $A, \theta_A$ to a final position and orientation $B, \theta_B$ when $\gamma \neq k 2\pi$. In contrast, Figure~\ref{fig:biarcAlphaColl} demonstrates how different biarcs, corresponding to different choices of $J$, can be utilized to move from an initial position and orientation $A, \theta_A$ to a final position with the same orientation $B, \theta_A$.'

\section{Choosing the joint location for the initial planning}
\label{sec:planJ}

Next, we shall delve into the process of selecting the optimal $J$. To commence, we will introduce the following additional definitions:
\begin{itemize}
\item $l_A=|\stackrel{\frown}{AJ}|$ denotes the length of the arc from point $A$ to point $J$. 
\item $l_B=|\stackrel{\frown}{JB}|$ signifies the length of the arc from point $J$ to point $B$.
\item $l$  represents the total length of the biarc, i.e., $l = l_A + l_B$. 
\item $r_A$  is defined as the signed radius of the arc from point $A$ to point $J$. A positive radius corresponds to counterclockwise rotation, while a negative radius corresponds to clockwise rotation.
\item $r_B$  is the signed radius of the arc from point $J$ to point $B$. A positive radius indicates counterclockwise rotation, and a negative radius indicates clockwise rotation. 
\item $k_A$ is defined as the signed curvature of the arc from point $A$ to point $J$, calculated as $k_A = \frac{1}{r_A}$.
\item $k_B$ represents the signed curvature of the arc from point $J$ to point $B$, determined by the equation $k_B = \frac{1}{r_B}$.
\end{itemize}

Minimizing the curvature discontinuity at the joint point is important in order to mitigate the acceleration demands placed on the robot. Nevertheless, when determining the optimal placement of $J$ to achieve reduced curvature discontinuity, it is essential to concurrently consider the path length. Notably, choices of $J$ that yield decreased curvature discontinuity may inadvertently result in excessively lengthy paths. The ensuing analysis will comprehensively investigate the disparities in curvature and path lengths associated with different $J$ selections, and elucidate the underlying rationale for the choice of $J$ adopted in this study.

In Section~\ref{sec:chooseJnondeg}, we will scrutinize the non-degenerate scenario where $\gamma \neq k 2\pi$. Subsequently, in Section~\ref{sec:chooseJdeg}, we will delve into the analysis of the degenerate case, characterized by $\gamma = k 2\pi$.

\subsection{Choosing the joint location for the initial planning, non degenerate case}
\label{sec:chooseJnondeg}

First, let us assume $\gamma \neq k 2\pi$. Let us introduce the following additional definitions (as depicted in Figure~\ref{fig:biarcDefs}):
\begin{itemize}
\item $\alpha$ is defined as the angle measured from $CM$ to $CJ$.
\item $-\pi \leq \delta_A < \pi$ represents the angle measured from the segment $AJ$ to the orientation of the robot at point $A$.
\item $-\pi \leq \delta_B < \pi$ denotes the angle measured from the segment $JB$ to the orientation of the robot at point $B$.
\item $\zeta$ is established as the angle measured from $AC$ to the orientation of the robot at point $A$. This angle is equivalent to the angle measured from $BC$ to the orientation of the robot at point $B$ (as described in~\cite{vsir2006approximating}).

\item $H_A$ is defined as the point of intersection with the circle of the line passing through $A$ with direction equal to the orientation of the robot in $A$. Notably, $H_A$ coincides with $A$ when $\zeta=\frac{\pi}{2}+k\pi$, where $k \in \mathbb{Z}$, that is when the orientation is tangential to the circle.
\item $H_B$ is similarly defined as the point of intersection with the circle of the line passing through $B$ with direction equal to the orientation of the robot in $B$. Again $H_B=B$ for $\zeta=\frac{\pi}{2}+k\pi$, $k \in \mathbb{Z}$, that is when the orientation is tangential to the circle.
\item $\phi_M$ is used to denote $\frac{\phi_A+\phi_B}{2}$.
\item $\alpha_{H_A}$ is designated as the angle from $CM$ to $CH_A$. It can be established that $\alpha_{H_A}=\phi_M+\phi_A$.
\item $\alpha_{H_B}$ is defined as the angle from $CM$ to $CH_B$. It can be derived that $\alpha_{H_B}=\phi_M+\phi_B$.
\end{itemize}

Given these definitions it can be demonstrated~\cite{su2014computational} that:
\begin{eqnarray}
r_A&=&\frac{|AB|}{2\sin\left( \frac{\gamma}{2} \right)}\frac{\sin\left(\frac{\alpha}{2}+\frac{\gamma}{4}\right)}{\sin\left(\frac{\alpha}{2}-\frac{\alpha_{H_A}}{2}\right)  }\\
r_B&=&\frac{|AB|}{2\sin\left( \frac{\gamma}{2} \right)}\frac{\sin\left(\frac{\alpha}{2}-\frac{\gamma}{4}\right)}{\sin\left(\frac{\alpha}{2} - \frac{\alpha_{H_B}}{2} \right) }\\
k_B-k_A&=&\frac{2 \sin\left( \frac{\gamma}{2}\right)  }{|AB|}\frac{\cos\left( \phi_B\right) - \cos\left(\phi_A\right)}{\cos\left(\frac{\gamma}{2} \right) - \cos\left( \alpha \right)  }\label{eq:kdiff}
\end{eqnarray}

\begin{figure}
\centering
\centering
\subfloat[][]{%
\label{fig:kAlpha}%
\includegraphics[width=\linewidth]{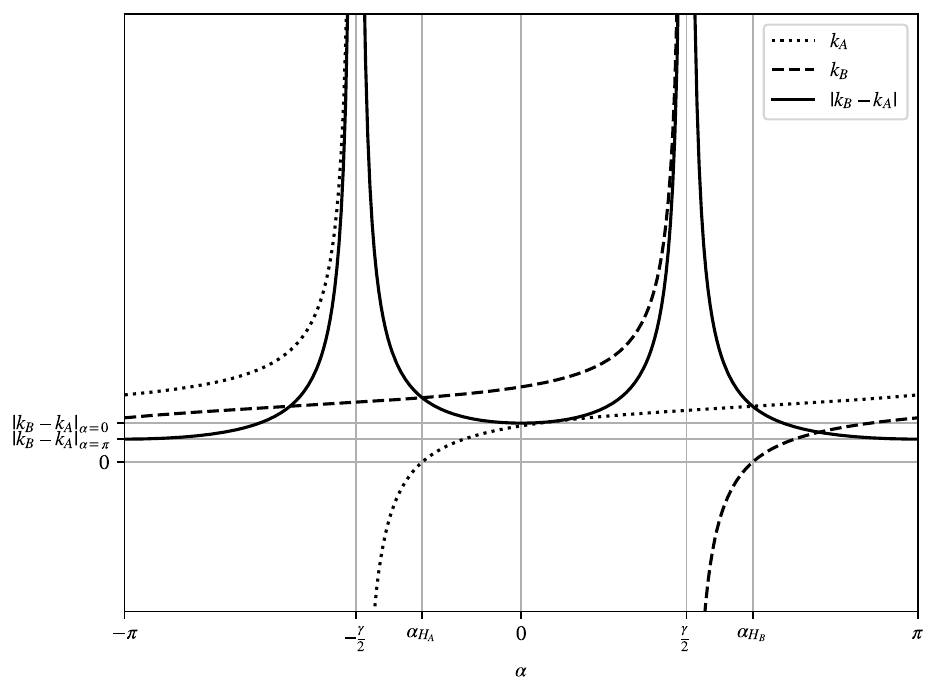}  }\\%
\hspace{8pt}%
\subfloat[][]{%
\label{fig:lengthAlpha}%
\includegraphics[width=\linewidth]{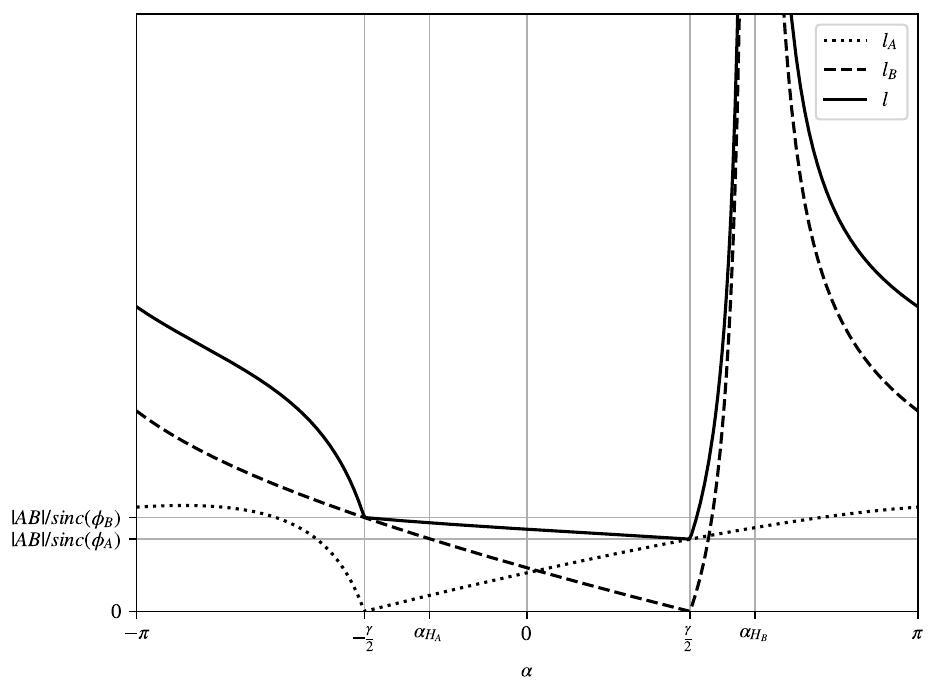}  }
  \caption{Curvature and path length obtained varying $\alpha$ for a biarc with $\phi_A=-\frac{\pi}{3}$ and $\phi_B=\frac{\pi}{2}$. Panel (a) reports the curvature of each arc and the absolute value of their difference. Panel (b) reports the path length for each arc and the total length.}
  \label{fig:krdiff}
\end{figure}

Upon examining Eq.~\ref{eq:kdiff}, it becomes evident that $|k_B-k_A|$ exhibits local minima at $\alpha=k \pi$, where $k\in \mathbb{Z}$. Specifically, the minimum occurs at $\alpha=k2\pi$ when $\cos\left( \frac{\gamma}{2} \right)<0$, and at $\alpha=\pi+k2\pi$ when $\cos\left( \frac{\gamma}{2} \right)>0$. In cases where $\gamma=\pi+k 2\pi$ (i.e.\, when $\cos\left( \frac{\gamma}{2} \right)=0$), both points yield the same absolute difference in curvature. As an example, Fig.~\ref{fig:kAlpha} illustrates the variation in curvature while adjusting $\alpha$, with initial orientations set to $\phi_A=-\frac{\pi}{3}$ and $\phi_B=\frac{\pi}{2}$.

Let us now proceed to analyze the path length by applying the law of sines:
\begin{eqnarray}
  |AJ|&=& |AB|\left|\frac{\sin\left(\frac{\alpha}{2}+\frac{\gamma}{4} \right)}{\sin\left( \frac{\gamma}{2} \right)}\right| \\
  |JB|&=& |AB|\left|\frac{\sin\left(\frac{\alpha}{2}-\frac{\gamma}{4} \right)}{\sin\left( \frac{\gamma}{2} \right)}\right|     
\end{eqnarray}

Denoting by $\sinc(x)$ the unnormalized sinc function, we can write:
\begin{eqnarray}
  l_A&=\frac{|AJ|}{\sinc(\delta_A)}\\
  l_B&=\frac{|JB|}{\sinc(\delta_B)}
\end{eqnarray}

If $\cos\left(\zeta\right)=0$ (which holds for $\cos(\phi_A+\phi_B)=1$) then
\begin{eqnarray}
  l&=&
  \begin{cases}
    \frac{|AB|}{\sinc\left(\frac{\gamma}{2}\right)} & \cos\left(\alpha \right)\geq \cos\left(\frac{\gamma}{2}\right)\\
    \frac{|AB| \frac{\gamma}{2}+\pi}{ \sin\left(\frac{\gamma}{2}\right)} & \cos\left(\alpha \right)<\cos\left(\frac{\gamma}{2}\right)
  \end{cases}.
  \label{eq:phiBecomesArc}
\end{eqnarray}
This reveals that the path length exhibits only two possible values as $\alpha$ varies. Moreover, the minimum path length is achieved for all values of $\alpha$ that fulfill the condition $\cos(\alpha) \geq \cos\left(\frac{\gamma}{2}\right)$.

If $\cos\left(\zeta\right)<0$, for $\alpha=\alpha_{H_A}$, $l_A$ (and consequently, $l$) becomes infinite. It can be demonstrated that $\cos\left(\zeta\right)<0$ is synonymous with $\cos(\phi_B) > \cos(\phi_A)$. Likewise, if $\cos\left(\zeta\right)>0$ (i.e., when $\cos(\phi_B) < \cos(\phi_A)$), we encounter infinite values for $l_B$ (and, consequently, $l$) when $\alpha=\alpha_{H_B}$. In the case where $\alpha=-\frac{\gamma}{2} + k 2\pi$, corresponding to $J=A$, the path length is given by $l = \frac{|AB|}{\sinc(\phi_B)}$. When $\alpha=\frac{\gamma}{2} + k 2\pi$, indicating the choice of $J=B$, the path length becomes $l= \frac{|AB|}{\sinc(\phi_A)}$.

The path length is a positive function of $\alpha$ and exhibits a single discontinuity. It tends to infinity at $\alpha = \alpha_{H_A} + k2\pi$ (when $\cos(\phi_B) > \cos(\phi_A)$) or at $\alpha = \alpha_{H_B} + k2\pi$ (when $\cos(\phi_B) > \cos(\phi_A)$). In all other cases, the path length remains finite and reaches a minimum at $\alpha = -\frac{\gamma}{2} + k2\pi$ (when $\cos(\phi_B) > \cos(\phi_A)$) or at $\alpha = \frac{\gamma}{2} + k2\pi$ (when $\cos(\phi_B) < \cos(\phi_A)$). Fig.~\ref{fig:lengthAlpha} shows an example of path lengths obtained  for $\phi_A=-\frac{\pi}{3}, \phi_B=\frac{\pi}{2}$.
 
Let us shift our attention to the path length for the two specific values of $\alpha$ that correspond to local minima in the curvature difference: $\alpha=0$ and $\alpha=\pi$. We will refer to the path length for $\alpha=0$ as $l_{\alpha=0}$ and to the path length for $\alpha=\pi$ as $l_{\alpha=\pi}$. Numerical analysis reveals that, for any given biarc, $l_{\alpha=0} < \frac{4}{3} l_{\alpha=\pi}$, and that $l_{\alpha=0}$ is smaller than $l_{\alpha=\pi}$  for all biarcs with initial and final orientations in the range of $-\frac{40}{41}\pi \leq \phi_A, \phi_B \leq \frac{40}{41}\pi$. In contrast, $l_{\alpha=0}$ exceeds $l_{\alpha=\pi}$ only in situations where $l_{\alpha=\pi}$ is already substantially large. To be specific, $l_{\alpha=\pi}$ surpasses $80|AB|$ in all cases where $l_{\alpha=0}$ is greater than $l_{\alpha=\pi}$. For this reason, although the minimum difference in curvature is achieved when $\alpha=\pi+k2\pi$ when $\cos\left( \frac{\gamma}{2} \right)>0$, in the following, we consistently choose $\alpha=0$ for any biarc. In other terms, we choose the point $J$ so that it lies on the segment bisector of $AB$, on the right side of $AB$ (facing B) when $\phi_B > \phi_A$ on the left when $\phi_B < \phi_A$, and on $AB$ itself when $\phi_A=\phi_B$.

This choice, commonly known as the 'equal chord' biarc in CAD/CAM applications~\cite{meek1992approximation,schonherr1993smooth,wang1993orientation}, results in the following set of equalities:
\begin{eqnarray}
  J_{\alpha=0}&=& M- \frac{|AB|}{2} \bm{v}\tan\left( \frac{\gamma}{4}\right) \label{eq:Jeq}\\
  |AJ|_{\alpha=0}&=&|JB|_{\alpha=0}=\frac{|AB|}{2 \cos\left(\frac{\gamma}{4} \right)}\label{eq:bisectChord}\\
  l_{A_{\alpha=0}}&=&\frac{|AJ|_{\alpha=0}}{\sinc\left( \frac{\alpha_{H_A}}{2}  \right) }\label{eq:bisectLa}\\
  l_{B_{\alpha=0}}&=&\frac{|JB|_{\alpha=0}}{\sinc\left( \frac{\alpha_{H_B}}{2}  \right) }\label{eq:bisectLb}\\
  k_{A_{\alpha=0}}&=&-\frac{2 \left( \sin(\phi_M)+\sin(\phi_A) \right)}{|AB|} \label{eq:bisectKa}\\
  k_{B_{\alpha=0}}&=&\frac{2 \left( \sin(\phi_M)+\sin(\phi_B) \right)}{|AB|} \label{eq:bisectKb}\\
  k_{B_{\alpha=0}}-k_{A_{\alpha=0}}&=&\frac{8\sin(\phi_M)\cos^2\left( \frac{\gamma}{4}\right) }{|AB|} \label{eq:bisectKd}
\end{eqnarray}

\subsection{Choosing the joint location for the initial planning,  degenerate case}
\label{sec:chooseJdeg}

Now, let us consider the degenerate case, in which the initial and final orientation are identical ($\gamma = k 2\pi$), resulting in the locus of $J$ forming a line.
Let us parameterize $J$ in terms of $\alpha \in \mathbb{R}$, with $J$ defined as
\begin{eqnarray}
  J&=& A\left( \frac{1}{2}-\alpha \right)+ B\left( \frac{1}{2}+\alpha \right)
\end{eqnarray}
we thus have $J=A$ for $\alpha=-\frac{1}{2}$ and $J=B$ for $\alpha=\frac{1}{2}$.

Let us define $\phi=\phi_A=\phi_B$ and  $\bar{\phi}=\pi - \phi_A +k2\pi$ with $k$ such that $-\pi \leq \bar{\phi} < \pi$.

The radii of curvature are
\begin{eqnarray}
  r_A&=&|AB|\frac{\alpha+\frac{1}{2}}{-2 \sin\left( \phi \right) }\\
  r_B&=&|AB|\frac{\alpha-\frac{1}{2}}{-2 \sin\left( \phi \right) }
\end{eqnarray}
and the curvature difference is
\begin{eqnarray}
  k_B-k_A&=&\frac{2\sin(\phi)}{|AB| \left(\frac{1}{4}- \alpha^2\right) }
\end{eqnarray}

If $\phi=k\pi$, the radii become infinite, and the curvatures are zero. Consequently, the curvature difference remains zero for any $\alpha$.
If $\phi \neq k\pi$, for $\alpha=-\frac{1}{2}$ (i.e., $J=A$), the curvature $k_A$ becomes infinite, and for $\alpha=\frac{1}{2}$ (i.e., $J=B$), the curvature $k_B$ becomes infinite. Thus, the curvature difference is infinite for either of these values of $\alpha$.
Conversely, the curvature difference tends to zero as $|\alpha|$ approaches infinity. It is worth noting that the curvature difference exhibits a local minimum at $\alpha=0$.

\begin{figure}
\centering
\centering
\subfloat[][]{%
\label{fig:zkAlpha}%
\includegraphics[width=\linewidth]{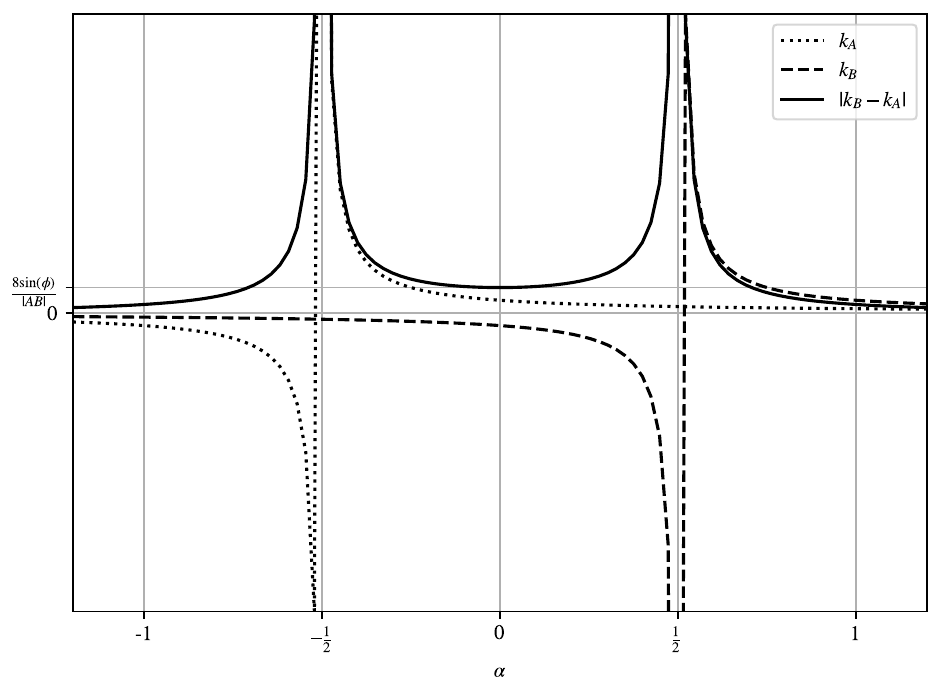}  }\\%
\hspace{8pt}%
\subfloat[][]{%
\label{fig:zlengthAlpha}%
\includegraphics[width=\linewidth]{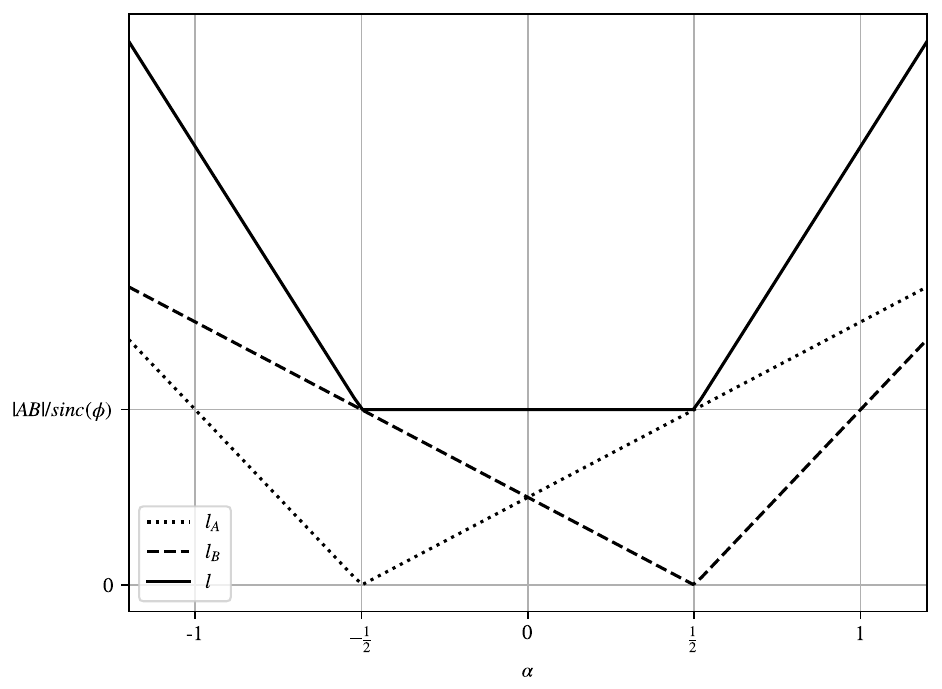}  }
  \caption{Curvature and path length obtained varying $\alpha$ for a biarc with $\phi_A=\phi_B=-\frac{\pi}{3}$. Panel (a) reports the curvature of each arc and the absolute value of their difference. Panel (b) reports the path length for each arc and the total length.}
  \label{fig:zalphaKL}
\end{figure}

In the degenerate case, the lengths of the two arcs are given by:
\begin{eqnarray} 
l_A&=&  \begin{cases}
          |AB|\frac{ -\frac{1}{2}-\alpha}{\sinc\left( \bar{\phi} \right)} &   \alpha<-\frac{1}{2} \land  \phi\neq 0 \\
          |AB|\frac{ \frac{1}{2}+\alpha}{\sinc\left( \phi \right)}  &   \alpha\geq-\frac{1}{2} \land  \phi\neq \pi\\
          \infty & \text{otherwise}
      \end{cases}\\
l_B&=&  \begin{cases}
          |AB|\frac{ \frac{1}{2}-\alpha}{\sinc\left( \phi \right)} &   \alpha\leq\frac{1}{2} \land \phi\neq \pi  \\
          |AB|\frac{ \alpha-\frac{1}{2}}{\sinc\left( \bar{\phi} \right)}  &   \alpha>\frac{1}{2}  \land \phi\neq 0 \\
          \infty & \text{otherwise}
        \end{cases}
\end{eqnarray}
therefore the total length is:
\begin{eqnarray*}%
  l&=& \begin{cases}
         \frac{|AB|}{\sinc\left( \phi \right)} &  \left| \alpha \right| \leq \frac{1}{2} \land \phi\neq \pi \\
         |AB| \frac{ \pi  \left( |\alpha|-\frac{1}{2}  \right) + |\phi| }{\left|\sin\left( \phi \right) \right|}&  \left| \alpha \right| > \frac{1}{2} \land \phi\neq k\pi, k\in \mathbb{Z}\\
         \infty & \text{otherwise}
     \end{cases}
\end{eqnarray*}
and clearly for $|\alpha|\rightarrow \infty$ we have $l \rightarrow \infty$. An example illustrating how the path characteristics vary by changing $\alpha$ for $\phi=-\frac{\pi}{3}$ can be seen in Fig.~\ref{fig:zalphaKL}.

Once again, we select $\alpha=0$ due to it being a local minimum for the curvature difference while not resulting in an excessively long path. Also with this choice, point $J$ lies on the bisector of $AB$. Furthermore, Eq.~\ref{eq:Jeq} to \ref{eq:bisectKd} remain applicable. In particular, these equations for $\phi=\phi_A=\phi_B$ simplify to:
\begin{eqnarray}
  J_{\alpha=0}&=&M \label{eq:jToMidpoint}\\
  |AJ|_{\alpha=0}=|JB|_{\alpha=0}&=&\frac{|AB|}{2}\\
  l_{A_{\alpha=0}}=l_{B_{\alpha=0}}&=&\frac{|AB|}{2\sinc(\phi)} \label{eq:zbisectLa}\\
  k_{A_{\alpha=0}}&=&-\frac{4\sin(\phi)}{|AB|}\\
  k_{B_{\alpha=0}}&=& \frac{4\sin(\phi)}{|AB|}\\
  k_{B_{\alpha=0}}-k_{A_{\alpha=0}}&=& \frac{8\sin(\phi)}{|AB|} \label{eq:zbisectKd}
\end{eqnarray}

\section{Choosing the joint location for replanning}
\label{sec:replanJ}

In most mobile robot applications, online replanning is a necessity due to various factors, including non-static obstacles in the environment and imperfections in the robot's trajectory caused by sensing noise, slippage, terrain irregularities, and more. As a result, we assume periodic replanning from the robot's current pose to the goal pose.

During replanning, it is highly desirable to keep the trajectory close to a previous plan. Drastic changes to the robot's trajectory at each time step can lead to jerky movements. Therefore, the plan generated in the previous time step should serve as a reference for constructing the new plan.

\begin{figure}
\centering
  \includegraphics[width=\linewidth]{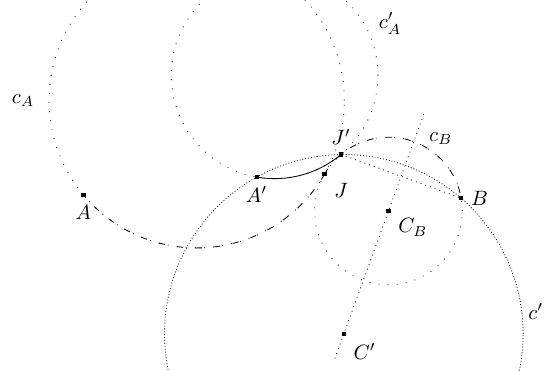}  
  \caption{Replanning heuristic example. The previously planned path is represented by $AJB$. $A'$ indicates the current position. The new point $J'$ is chosen to ensure an overlap between the segments $J'B$ and $JB$.}
  \label{fig:feedback}
\end{figure}

At time $t_i$, a trajectory comprising a sequence of biarcs is generated. The first biarc, denoted as $b$, consists of two arcs, $\stackrel{\frown}{AJ}$ and $\stackrel{\frown}{JB}$, which guide the robot from its pose at $(A,\theta_A)$ to its pose at $(B,\theta_B)$. At time $t_{i+1}$, the robot is at pose $(A',\theta_A')$. The objective is to plan a new biarc, referred to as $b'$, from $(A',\theta_A')$ to $(B,\theta_B)$ in a manner that maintains proximity to the previous biarc $b$. Determining the closest biarc based on the squared distance between the two trajectories can be achieved through classical optimization techniques. However, this approach is computationally expensive. Therefore, we employ a heuristic method, as illustrated in Fig.~\ref{fig:feedback}.

Let $c_A$ represent the circle of which arc $\stackrel{\frown}{AJ}$ is a part. If $AJ$ is a straight segment (with zero curvature), then $c_A$ denotes the line that contains $AJ$. Similarly, let $c_B$ denote the circle of which arc $\stackrel{\frown}{JB}$ is a part or, in the case of $JB$ being a straight segment, the line defined by $J$ and $B$. In instances where $c_B$ represents a circle, refer to the center of $c_B$ as $C_B$.

Let us consider a potential candidate for a biarc $b'$, designed to resemble the previous biarc $b$.  This candidate biarc $b'$ consists of arcs $\stackrel{\frown}{A'J'}$ and $\stackrel{\frown}{J'B}$, with the condition that $J'$ lies on $c_B$ and $J' \neq B$. This approach represents a particular case of a heuristic introduced in the field of curve approximation by~\cite{vsir2006approximating}. With this choice, the chosen biarc and the original biarc overlap in their second arc. In Fig.~\ref{fig:feedback}, the previously planned biarc $b$ is represented by a dashed line. Biarc $b'$ consists of arc $A'J'$, illustrated by a solid line, and arc $J'B$, which is a segment of the original arc $JB$.

Let $c'$ represent the locus of points where $J'$ can be situated. As explained in Section~\ref{sec:biarcDef}, if $\theta_A' \neq \theta_B + k2\pi$, then $c'$ takes the form of a circle with the center $C'$, defined in Eq.~\ref{eq:cpos}, passing through point $B$. However, if $\theta_A' = \theta_B + k2\pi$, then $c'$ is represented by the line $A'B$. With our definition of a similar biarc, $J'$ must lie within the intersection of $c_B$ and $c'$. This intersection always includes at least one point, which is point $B$. In the case where both $c_B$ and $c'$ are not degenerate and $C'\neq C_B$, the position of $J'$ can be calculated as follows:
\begin{eqnarray}
  \bm{w}&=&\left[\begin{matrix} w_x\\ w_y \end{matrix} \right]=\frac{C_B-C'}{|C_BC'|}\\
  a_c&=&w_x^2-w_y^2\\
  a_s&=&2 w_xw_y\\%
  J'&=&\left[\begin{matrix} a_c & a_s\\ a_s & -a_c  \end{matrix} \right] B +\left[\begin{matrix} 1- a_c & -a_s\\ -a_s & 1+a_c  \end{matrix} \right] C'
\end{eqnarray}
The example depicted in Fig.~\ref{fig:feedback} illustrates this situation.
If both $c_B$ and $c'$ are circular and concentric, then $c_B$ coincides with $c'$, making any point $J' \in c'$ a valid choice. In this scenario, we select $J'$ corresponding to $\alpha=0$.

\begin{figure}%
\centering
\subfloat[][]{%
\label{fig:tracelincur}%
    \includegraphics[width=0.46\linewidth]{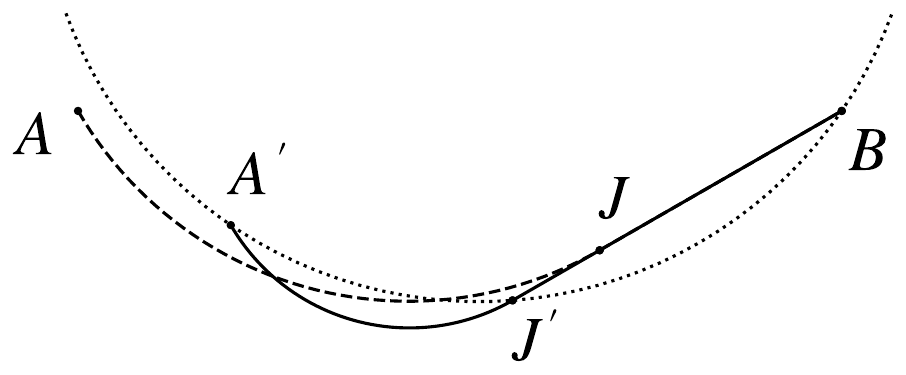}  }%
\hspace{8pt}%
\subfloat[][]{%
\label{fig:tracecurlin}%
\includegraphics[width=0.46\linewidth]{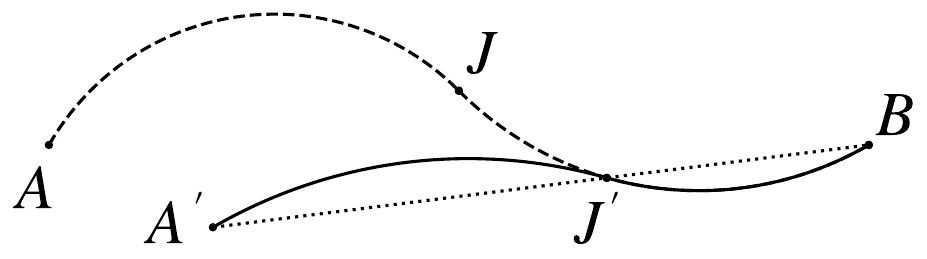}  }
\caption{Examples of replanning for a degenerate reference arc (left panel) and a degenerate joint point locus (right panel). The non-degenerate case can be seen in Fig.~\ref{fig:feedback}.}%
\label{fig:lineReflections}%
\end{figure}

If $c_B$ takes the form of a line, while $c'$ is a circle, then $J'$ corresponds to the reflection of point $B$ over an axis that passes through $C'$ and is perpendicular to the line $JB$. This operation can be computed efficiently as
\begin{eqnarray}
  J&=&B+ 2 \bm{u_{b}}^T (C' - B) \bm{u_{b}}
\end{eqnarray}
where $\bm{u_{b}}$ represents the unit vector in the direction of $JB$. An illustrative example can be found in Figure~\ref{fig:tracelincur}.

In a similar fashion, if $c_B$ is a circle and $c'$ is a line, then $J'$ can be determined as the reflection of point $B$ over an axis that passes through $C_B$ and is orthogonal to the line $A'B$. The computation for finding $J$ in this case is also straightforward:
\begin{eqnarray}
  J&=&B+ 2 {\bm{{u'}}^T (C_b - B) \bm{ {u'} }}
\end{eqnarray}
where $\bm{u'}$ is the unit vector in the direction $A'B$. An example is presented in Figure~\ref{fig:tracecurlin}.

If both $c_B$ and $c'$ are lines, their only point of intersection is $B$, rendering the existence of $b'$ impossible. In this case, $J$ is determined using Eq.~\ref{eq:jToMidpoint}.

\begin{figure}
\centering

  \includegraphics[width=\linewidth]{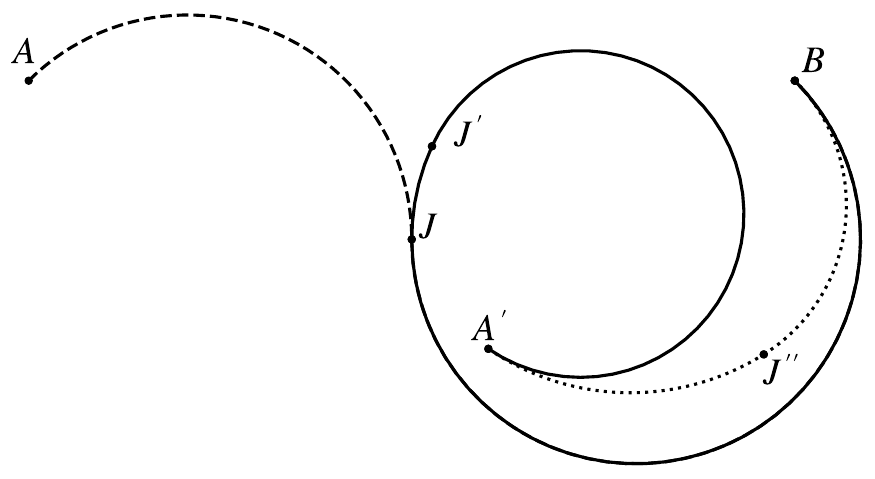}  
  \caption{An example for which the heuristic provides an undesirable biarc. $(\stackrel{\frown}{AJ}$,$\stackrel{\frown}{JB})$ is the biarc planned at the previous time step. $(\stackrel{\frown}{A'J'}$,$\stackrel{\frown}{J'B})$ is the biarc obtained using the heuristic. $(\stackrel{\frown}{A'J''}$,$\stackrel{\frown}{J''B})$ is the biarc from $A'$ to $B$ obtained for $\alpha=0$.}
  \label{fig:traceLong}
\end{figure}

In most cases, selecting the biarc $b'$ using the heuristic defined above results in good biarcs that resemble the original trajectory $b$. However, there are instances where the obtained biarc is excessively long or contains arcs with very high curvature, as illustrated in Fig~\ref{fig:traceLong}. If the biarc $b = (\stackrel{\frown}{AJ}, \stackrel{\frown}{JB})$ (depicted by a dashed line) is chosen as a reference for generating a biarc from $A'$ to $B$, the solid line biarc $b' = (\stackrel{\frown}{A'J'}, \stackrel{\frown}{J'B})$ is obtained. Conversely, when a biarc is planned from $A'$ to $B$ using $\alpha=0$, akin to the initial planning, the much more favorable dotted biarc $b'' = (\stackrel{\frown}{A'J''}, \stackrel{\frown}{J''B})$ is obtained.

Therefore, during the replanning process, a comparative analysis is conducted between the length and curvature difference of biarc $b'$ and those of biarc $b''$. The length of $b''$ can be calculated using Equations \ref{eq:bisectLa} and \ref{eq:bisectLb} when $\gamma \neq k 2\pi$, or Equation \ref{eq:zbisectLa} when $\gamma = k 2\pi$. Similarly, the curvature difference of $b''$ is determined using Equation \ref{eq:bisectKd} when $\gamma \neq k 2\pi$, and Equation \ref{eq:zbisectKd} when $\gamma = k 2\pi$. To decide between choosing $b'$ and $b''$, we introduce two parameters, $\eta_l$ and $\eta_k$, both greater than 1. The selection criterion is based on the conditions $l' > \eta_l \cdot l''$ or $|k_B' - k_A'| > \eta_k \cdot |k_B'' - k_A''|$. In our experimental evaluations, we determined that setting $\eta_l = \eta_k = 2$ yields favorable results.

\section{Collision detection}
\label{sec:collisionDetection}

Biarcs were selected as motion primitives for a specific reason—this type of trajectory lends itself to efficient collision detection. When a generic curve is employed as a motion primitive, it necessitates sampling the robot's position along the trajectory and subsequently calculating whether a collision occurs at each sampled point. As a result, the computational time scales roughly in proportion to the path length and the sampling resolution.

The primary benefit of using biarcs as motion primitives is the elimination of this sampling requirement. Collision detection with obstacles can be carried out through closed-form computations.

Specifically, collision detection for a biarc trajectory can be executed by checking for collisions in each of its two constituent arcs. Let's consider a single arc commencing at position $A_i$ with orientation $\theta_{A_i}$ and terminating at position $B_i$ with orientation $\theta_{B_i}$. In the case of a non-degenerate arc, we define the center of rotation as $C_i$ and the spanned angle as $\theta_{d_i} = \theta_{B_i} - \theta_{A_i}$.

For the subsequent discussions, we assume that a convex polygonal hitbox is utilized to represent the robot. If the hitbox is concave, it can be subdivided into convex hitboxes, and the collision check can be performed for each of these components.

This hitbox can be defined by its vertices as $H_j$, where $0 \leq j < N_h$ when the robot is located at position $A_i$. We can further define $H_{N_h} = H_0$, and denote each edge as $H_jH_{j+1}$.

We make the assumption that obstacles in the environment are static. This assumption is supported by the characteristics of our setup, where collision detection can be performed within a single LIDAR scan (taking 50ms), and the robot operates at relatively low speeds (6 km/h).

We investigate collision scenarios involving the movement of the robot's hitbox along an arc trajectory and its interaction with two types of obstacles. Section~\ref{sec:coldetPoint} focuses on point obstacles, which are a natural representation for LIDAR data. Section~\ref{sec:coldetLine} addresses collisions with line (and line segments) obstacles, serving as a foundation for assessing collision with polygonal obstacles, lane edges, walls, and similar structures.

It is worth noting that when the map is known in advance and static obstacles can be represented as polygonal objects, LIDAR data collected during robot navigation can be selectively omitted from collision detection if they fall inside or in close proximity to the edges of these obstacles. This optimization can significantly accelerate the collision evaluation process.

\subsection{Collision detection for point obstacles}
\label{sec:coldetPoint}

This section provides an efficient method for conducting collision detection between a convex polygonal hitbox in motion along an arc and a point.

Throughout the remainder of this section, point obstacles are represented by their coordinates, denoted as $P_k$, $1\leq i \leq N_p$. Two scenarios are distinguished: non-degenerate circular arc movements and straight-line movements.

\subsubsection{Collision detection for point obstacles, arc movement}
\label{sec:pointArc}

Collisions may occur either because a point is initially inside the hitbox at the start of the arc or because it enters the hitbox through one of its edges. Consider a path composed of a sequence of biarcs $b_0,\ldots,b_{i-1},b_{i},\ldots,b_n$. If no collision is detected for biarc $b_{i-1}$, then it is guaranteed that no point can be inside the hitbox at the beginning of biarc $b_i$. Therefore, the check for points inside the hitbox at the start of the arc only needs to be performed for the first arc of $b_0$. In our setup the initial point of the first arc of $b_0$ corresponds to the current robot position. While it is physically impossible for obstacles to be inside the robot itself, there may still be obstacles within the hitbox, necessitating the check for the initial position of the first arc.

Checking whether any of the obstacle points reside inside the hitbox can be done by projecting each point $P_k$ along the normals of the edges $H_jH_{j+1}$, and checking whether it lies on the half-plane belonging to the hitbox.

Now, we will focus on determining if a point $P_k$ enters the hitbox through the edge $H_jH_{j+1}$ while moving along the arc $\stackrel{\frown}{A_iB_i}$.
For this analysis, let us assume that the robot is halfway through its movement along the arc, and we establish a reference frame with the origin at the center of rotation $C_i$. In this context, we examine the line $l$ of which $H_jH_{j+1}$ is a segment.

A point $\bm{p}$ on this line can be parametrized as follows:
\begin{eqnarray}
  \bm{p}&=&  x \bm{u} +h \bm{u^\perp} \label{eq:lineParametrizationArc}
\end{eqnarray}
where $x, h \in \mathbb{R}$, $\bm{u} \in \mathbb{R}^{2x1}$ is a unitary vector parallel to the line $l$ and $\bm{u^\perp}\in \mathbb{R}^{2x1}$ is the perpendicular vector defined as
\begin{eqnarray}
  \bm{u^\perp}=\left[\begin{matrix}0 & -1\\1 & 0\end{matrix}\right]\bm{u}.
\end{eqnarray}

Let $x_j$ and $x_{j+1}$ represent the coordinates of $H_j$ and $H_{j+1}$, respectively. Now, let us examine the points on line $l$ where a point $P_k$ might enter the hitbox. If $\Vert P_k \Vert^2 - h_j^2 < 0$, there are no such points. If $\Vert P_k \Vert^2 - h_j^2 \geq 0$, then the only two possible (potentially coincident) entry points have coordinates $x_{k1},x_{k2}=\pm \sqrt{\Vert P_k \Vert - h_j^2 }$. Each of these points serves as the entry point if and only if it satisfies the following inequalities:

Let us denote by $x_j$ and $x_{j+1}$ the coordinates of $H_j$ and $H_{j+1}$, respectively.
Let us now consider the points of the line $l$ through which a point  $P_k$ may enter the hitbox.
If  $\Vert P_k \Vert^2 - h_j^2 <0$ then no such point exists. If $\Vert P_k \Vert^2 - h_j^2 \geq 0$ then the only two (possibly coincident) possible points of entrance have coordinates $x_{k1},x_{k2}=\pm\sqrt{\Vert P_k \Vert - h_j^2 }$. Each of such points is actually the point of entrance if and only if it satisfies the following inequalities:
\begin{eqnarray}
  x_{j} &\leq& x \leq x_{j+1} \\
  P_k^T  (x\bm{u} +h \bm{u^\perp}) &\geq& \Vert P_k \Vert^2 \cos\left( \frac{\phi_d}{2}\right).
\end{eqnarray}

\subsubsection{Collision detection for point obstacles, straight movement}
\label{sec:pointSegment}

Let us denote by $H'_j$ the vertices of the hitbox when the hitbox is at the arrival position $B_i$. Consider the convex hull of the points in the set:
\begin{equation}
H_{AB}=\{H_j : 0 \leq j \leq N_h\} \cup \{H'_j : 0 \leq j \leq N_h\}. \label{eq:convexHull}
\end{equation}

Collision can be detected by checking whether any of the obstacle points $P_k$ falls inside this convex hull. It is worth noting that for rectangular hitboxes, the computation is straightforward as the convex hull is also a rectangle.

\subsection{Collision detection for line obstacles}
\label{sec:coldetLine}

Let us now turn our attention to the collision between a hitbox edge $H_jH_{j+1}$ and an obstacle in the form of a line segment $S_lS_{l+1}$ or a line. We use $a$ to denote the line of which $H_jH_{j+1}$ is part of, and similarly, let line $b$ indicate the line of which $S_lS_{l+1}$ is part of. If the obstacle is an (infinite) line obstacle, then $b$ will denote the obstacle itself. As we did for point obstacles, we distinguish the non-degenerate case from the degenerate case.

\subsubsection{Collision detection for line obstacles, arcs movement}
\label{sec:lineArc}

Let us assume the radius of rotation is finite. Line $a$ rotates around the center of the arc $C_i$, while line $b$ is static. We consider a frame of reference with the origin at the center of rotation $C_i$. We apply a parametrization similar to the approach used in Section~\ref{sec:pointArc}. Here, a generic point $\bm{p_{l}}$ on line $l$ can be expressed as:
\begin{equation}
  \bm{p_{l}} = x_l \bm{u_{l}} + h_l \bm{u_{l}^\perp} \label{eq:lineParametrization}.
\end{equation}
In this equation, $x_l$ represents the displacement of the point along line $l$, $\bm{u_{l}}$ is a unitary vector parallel to line $l$, and it is chosen such that $h_l$ is non-negative and denotes the distance of the line from the origin. The perpendicular vector $\bm{u_{l}^\perp}$ is defined as:
\begin{equation}
  \bm{u^\perp_{l}} = \left[\begin{matrix} 0 & -1 \\ 1 & 0 \end{matrix}\right]\bm{u_{l}}.
\end{equation}
We establish a reference frame with its origin at point $C_i$. The rotation acting on line $a$ can be described by a matrix $\bm{R}$, defined as:
\begin{eqnarray}
  \bm{R}&=&  \left[\begin{matrix} 
          c_t &  -s_t \\
          s_t &  c_t
        \end{matrix}\right]
            \left[\begin{matrix}
          c_{\mu} &  -s_{\mu} \\
          s_{\mu} &  c_{\mu}
        \end{matrix}\right] \label{eq:rotationParametrization}
\end{eqnarray}
where
\begin{eqnarray}
  c_{\mu}&=&\bm{u_{b}}^T \bm{u_{a}} \label{eq:cmu}\\
  s_{\mu}&=& -{\bm{u_{b}^\perp}}^T \bm{u_{a}} = {\bm{u_{a}^\perp}}^T \bm{u_{b}} \label{eq:smu}\\
  c_t&=& \frac{1-t^2}{1+t^2}\\
  s_t&=& \frac{2t}{1+t^2}.
\end{eqnarray}
In other words, we express $\bm{R}$ as a composition of two rotations. First, with the right-hand matrix, we rotate $a$ to align it parallel to $b$. Then, with the left-hand matrix, we apply an additional rotation with an angle whose half tangent is $t$.

Let us define
\begin{enumerate}
\item $I_{a}$ as the set of rotations $t$ for which $H_jH_{j+1}$ intersect $b$.
\item $I_{b}$ as the set of rotation $t$ for which line $a$ intersect $S_lS_{l+1}$.
\item $I_{d_i}$ as the set of rotations $t$ taken when the robot performs the arc of interest.
\end{enumerate}

If the obstacle is a line segment $S_{l}S_{l+1}$, then collision with $H_jH_{j+1}$  occurs if and only if
\begin{eqnarray}
  I_{a} \cap I_{b} \cap I_{d_i} \neq \emptyset \label{eq:intersectionSegment}.
\end{eqnarray}
If the obstacle is a line $b$, the condition simplifies to
\begin{eqnarray}
  I_{a} \cap I_{d_i} \neq \emptyset  \label{eq:intersectionLine}.
\end{eqnarray}

The set $I_{d_i}$ is determined as follows. Let us define 
\begin{eqnarray}
  t_{d-}=\begin{cases}
           \tan\left(\frac{-\mu+\theta_{d_i}}{2}\right)&\text{if } \theta_{d_i}< 0\\
           \tan\left(\frac{-\mu}{2}\right)&\text{if } \theta_{d_i}\geq 0 \label{eq:tdminus}
         \end{cases}\\
  t_{d+}=\begin{cases}
           \tan\left(\frac{-\mu+\theta_{d_i}}{2}\right)&\text{if } \theta_{d_i}\geq 0\\
           \tan\left(\frac{-\mu}{2}\right)&\text{if } \theta_{d_i}< 0 \label{eq:tdplus}
         \end{cases}
\end{eqnarray}
where $\mu$ is the angle whose cosine is $c_{\mu}$ and whose sine is $s_{\mu}$ (see Eq.~\ref{eq:cmu} and Eq.~\ref{eq:smu}, respectively). With these definitions we can write:
\begin{eqnarray}
  I_{d_i}=\begin{cases}
            \left[ t_{d-}, t_{d+} \right]  &\text{if } t_{d-} \leq t_{d+} \\
            \left( -\infty, t_{d+} \right] \cup \left[t_{d-},+\infty \right)  &\text{if }  t_{d-} > t_{d+}
          \end{cases}\label{eq:idi}.
\end{eqnarray}

The determination of sets $I_{a}$ and $I_{b}$ is more involved.
Using the line parametrization from Eq.~\ref{eq:lineParametrization} and the rotation parametrization from Eq.\ref{eq:rotationParametrization}, for a given rotation $t$ where $t\neq 0$, the point of intersection between the rotated line $a$ and line $b$ is determined as follows:
\begin{eqnarray}
  x_a&=&\frac{\left(h_{a} + h_{b}\right) t^{2} +h_{b} - h_{a}}{2 t} \label{eq:xa}\\
  x_b&=&-\frac{\left(h_{a} + h_{b}\right) t^{2}+h_{a} - h_{b}}{2 t} \label{eq:xb}.
\end{eqnarray}
In the case of $t=0$, the lines either coincide and intersect at any point, or they are parallel and have no intersection.

When considering a coordinate $x_a$ such that $x_a^2 \geq h_b^2 - h_a^2$, two rotations, denoted as $t_{a+}$ and $t_{a-}$, result in this coordinate being an intersection. These rotations are calculated as follows:
\begin{eqnarray}
  t_{a+}&=&\frac{x_a + \sqrt{x_a^{2}+h_{a}^{2} - h_{b}^{2}}}{h_{a} + h_{b}}\label{eq:taplus}\\
  t_{a-}&=&\frac{x_a - \sqrt{x_a^{2}+ h_{a}^{2} - h_{b}^{2}}}{h_{a} + h_{b}}\label{eq:taminus}
\end{eqnarray}
Likewise, a coordinate $x_b$ with $x_b^2 \geq h_a^2 - h_b^2$ can be an intersection by performing the rotation:
\begin{eqnarray}
  t_{b+}&=&\frac{-x_b + \sqrt{ x_b^{2}+h_{b}^{2} - h_{a}^{2} }}{h_{a} + h_{b}}\label{eq:tbplus}\\
  t_{b-}&=&\frac{-x_b - \sqrt{ x_b^{2}+ h_{b}^{2} - h_{a}^{2} }}{h_{a} + h_{b}}\label{eq:tbminus}.
\end{eqnarray}

Let us denote the coordinates of the points $H_{j}$ and $H_{j+1}$ on $a$ according to Eq.~\ref{eq:lineParametrization} as
\begin{eqnarray}
  x_{a1}&=&\min\left\{ u_a^T H_{j},u_a^T H_{j+1}\right\}\\
  x_{a2}&=&\max\left\{u_a^T H_{j},u_a^T H_{j+1}\right\}
\end{eqnarray}
and denote the corresponding values of $t$, computed using Eq.~\ref{eq:taminus} and Eq.~\ref{eq:taplus} as $t_{a1-},t_{a1+},t_{a2-},t_{a2+}$.
In an analogous way, let us express the coordinates of $S_{l}$ and $S_{l+1}$ on $b$ as $x_{b1}$ and $x_{b2}$, with associated $t$ values $t_{b1-},t_{b1+},t_{b2-},t_{b2+}$.

Let us now consider how the intersection point moves over the lines when rotating $a$.
The first order derivatives of $x_a$ and $x_b$with respect to  $t$ are given by
\begin{eqnarray}
  \frac{d x_a}{dt}&=&\frac{\left(h_{a} + h_{b}\right) t^{2} + h_{a} - h_{b}}{2 t^{2}} \label{eq:dxadt}\\
  \frac{d x_b}{dt}&=&-\frac{\left( h_{a}+ h_{b}\right) t^{2}+  h_{b} - h_{a} }{2 t^{2}}\label{eq:dxbdt}
\end{eqnarray}
and the second order derivatives are
\begin{eqnarray}
  \frac{d^2 x_a}{dt^2}&=&\frac{d^2 x_b}{dt^2}=\frac{h_{b} - h_{a}}{t^{3}}\label{ex:dx2dt}.
\end{eqnarray}

From these derivatives we evince that the computation of $I_{a}$ and $I_{b}$ requires distinguishing three cases  based on the values of $h_a$ and $h_b$: when $h_a > h_b$, when $h_a = h_b$, and when $h_a < h_b$.

\begin{figure}%
\centering
\subfloat[][]{%
\label{fig:thagthb}%
    \includegraphics[width=\linewidth]{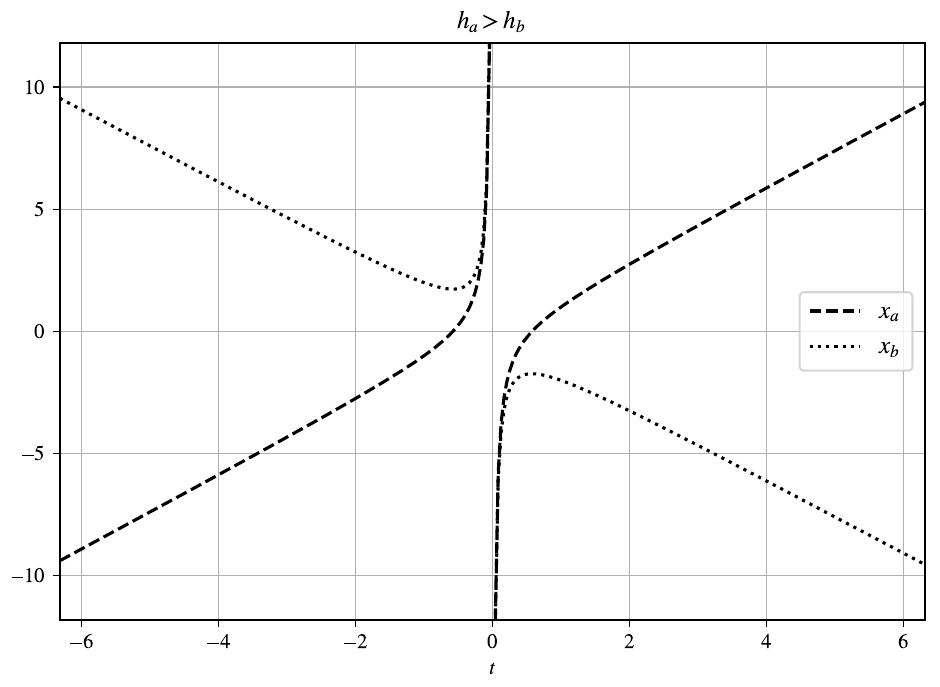}  }\\
\subfloat[][]{%
\label{fig:thaeqhb}%
    \includegraphics[width=\linewidth]{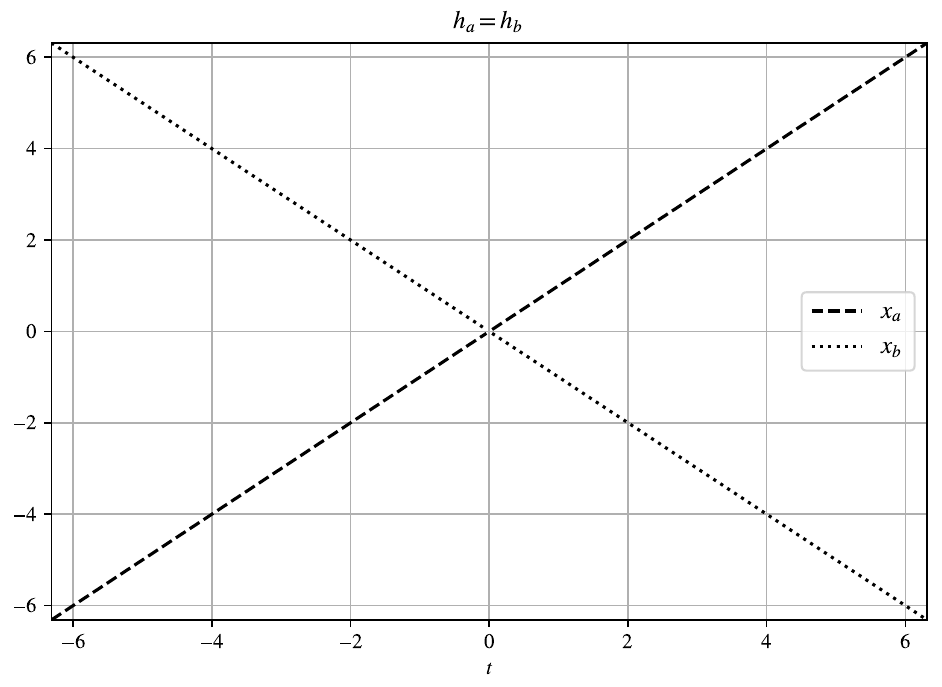}  }]\\
\subfloat[][]{%
\label{fig:thalthb}%
    \includegraphics[width=\linewidth]{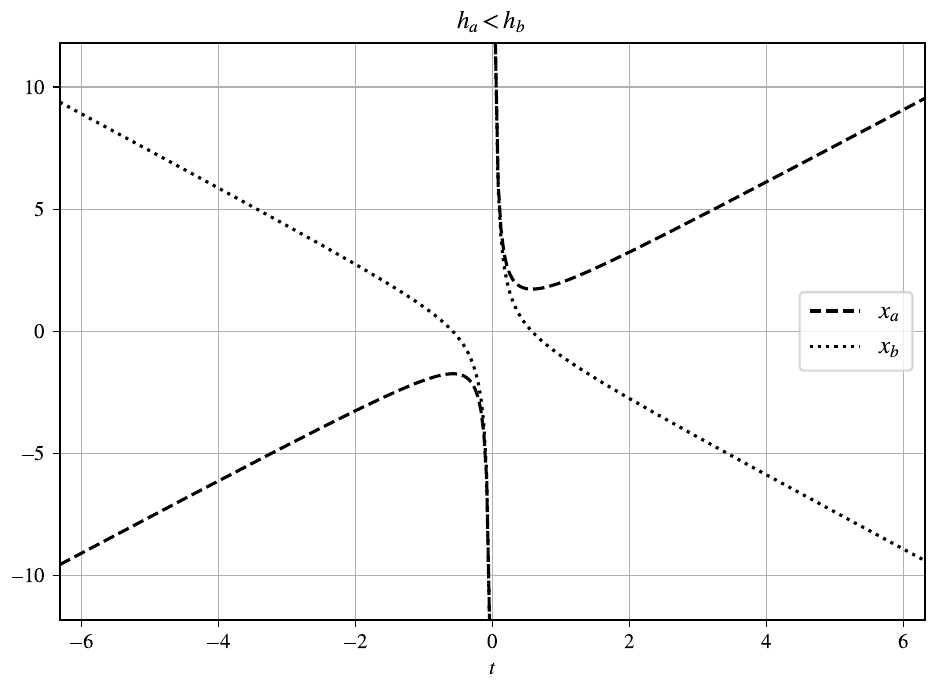}  }%
\caption{Example of $x_a$ and $x_b$ in terms of $t$ for $h_a=2>h_b=1$ (panel a), $h_a=h_b=1$ (panel b) and $h_a=1<h_b=2$ (panel c). The horizontal axis reports the value $t$. The vertical axis reports the values of $x_a$ (dashed curve) and $x_b$ (dotted curve).}%
\label{fig:thahbcurves}%
\end{figure}

In the case where $h_a > h_b$ (as shown in Fig.~\ref{fig:thagthb}), the following holds:
\begin{eqnarray}
  \frac{d x_a}{dt}&>& 0 \;\forall t\\
  \lim_{t \to -\infty} x_a &=& \lim_{t \to 0^+} x_a = -\infty \\
  \lim_{t \to 0^-} x_a &=& \lim_{t \to -\infty} x_a=+\infty .
\end{eqnarray}
Collision occurs inside the segment $H_{j}H_{j+1}$ for the following range of values:
\begin{eqnarray}
  I_{a}&=&[t_{a1-},t_{a2-}] \cup [t_{a1+},t_{a2+}] \label{eq:iahagthb}.
\end{eqnarray}
In the case of line $b$, we have the following limit values:
\begin{eqnarray}
  \lim_{t \to -\infty} x_b &=&\lim_{t \to 0^-} x_b = +\infty \\
  \lim_{t \to 0^+} x_b &=&\lim_{t \to +\infty} x_b = -\infty
\end{eqnarray}
The derivative of $x_b$ with respect to $t$ is zero at two points:
\begin{eqnarray}
  t_{bz-}&=-\sqrt{\frac{h_a-h_b}{h_a+h_b}}\\
  t_{bz+}&=\sqrt{\frac{h_a-h_b}{h_a+h_b}}
\end{eqnarray}
with
\begin{eqnarray}
  \left.\frac{d^2 x_b}{dt^2} \right\vert_{t=t_{bz+}}&<&0<\left.\frac{d^2 x_b}{dt^2} \right\vert_{t=t_{bz-}}.
\end{eqnarray}
At $t_{bz-}$, there is a local minimum with $x_{bz-} = b^{*}$, and at $t_{bz+}$, there is a local maximum with $x_{bz+} = -b^{*}$, where
\begin{eqnarray}
  b^{*}&=&\sqrt{h_a^2-h_b^2}
\end{eqnarray}
From this analysis, we can deduce that the intervals of rotations $t$ that result in a collision for segment $S_{l}S_{l+1}$ are as follows:
{\small
\begin{eqnarray}
  I_{b}=
  \begin{cases}
    [t_{b1-},t_{b2-}]\cup [t_{b2+},t_{b1+}] &\text{if } x_{b1}\leq x_{b2} < -b^{*}\\
    [t_{b1-},t_{b1+}]                  &\text{if } x_{b1}\leq -b^{*} \leq  x_{b2} < b^{*}  \\
    [t_{b2-},t_{b2+}]\cup[t_{b1-},t_{b1+}] &\text{if } x_{b1}\leq -b^{*}  \leq b^{*} \leq  x_{b2} \\
    \emptyset&\text{if } -b^{*} < x_{b1} \leq  x_{b2} <  b^{*} \\
    [t_{b2-},t_{b2+}]&\text{if } -b^{*} < x_{b1}<  b^{*} \leq  x_{b2} \\
    [t_{b2-},t_{b1-}]\cup [t_{b1+},t_{b2+}] &\text{if } b^{*}\leq x_{b1} \leq x_{b2}
  \end{cases} \label{eq:ibhagthb}.
\end{eqnarray}}

In the scenario where $h_a = h_b$, as depicted in Fig.~\ref{fig:thaeqhb}, the following observations can be made. At $t=0$, all points intersect because $a$ and $b$ overlap.
For $t \neq 0$:
\begin{eqnarray}
  x_a&=&-x_b= h_a t\\
  \frac{d x_a}{dt}&\geq& 0 \;\forall t\\
  \frac{d x_b}{dt}&\leq& 0 \;\forall t.
\end{eqnarray}
Consequently, for the case of $h_a=h_b$, the intervals of collisions are as follows:
\begin{eqnarray}
  I_{a}&=&\left[\frac{x_{a1}}{h_a},\frac{x_{a2}}{h_a}\right]\cup\{0\} \label{eq:iahaeqhb}\\
  I_{b}&=&\left[-\frac{x_{b2}}{h_a},-\frac{x_{b1}}{h_a}\right]\cup\{0\}\label{eq:ibhaeqhb}.
\end{eqnarray}

The remaining possibility $h_a<h_b$ is depicted in Fig.~\ref{fig:thalthb}. In this case the following holds:
\begin{eqnarray}
  \frac{d x_b}{dt} &< 0 \forall t\\
  \lim_{t \to -\infty} x_a  &=&   \lim_{t \to 0^-} x_a  = -\infty \\
  \lim_{t \to 0^+} x_a&=& \lim_{t \to +\infty} x_a = +\infty
\end{eqnarray}
and $\frac{d x_a}{dt}=0$ for
\begin{eqnarray}
  t_{az-}&=-\sqrt{\frac{h_b-h_a}{h_a+h_b}}\\
  t_{az+}&=\sqrt{\frac{h_b-h_a}{h_a+h_b}}
\end{eqnarray}
with
\begin{eqnarray}
  \left.\frac{d^2 x_a}{dt^2} \right\vert_{t=t_{az-}}&<&0<\left.\frac{d^2 x_a}{dt^2} \right\vert_{t=t_{az+}}.
\end{eqnarray}
We thus have a local maximum at $t_{az-}$ and  a local minimum at $t_{az+}$, with $x_{az-}=-a^{*},x_{az+}=a^{*}$ where
\begin{eqnarray}
  a^{*}&=&\sqrt{h_b^2-h_a^2}.
\end{eqnarray}  
The intervals of collision for segment $H_{j}H_{j+1}$ are
{\small
  \begin{eqnarray}
    I_{a}=
  \begin{cases}
    [t_{a1-},t_{a2-}]\cup [t_{a2+},t_{a1+}] &\text{if } x_{a1}\leq x_{a2} \leq -a^{*}\\
    [t_{a1-},t_{a1+}]                  &\text{if } x_{a1}\leq  -a^{*} < x_{a2} < a^{*} \\
    [t_{a1-},t_{a1+}]\cup[t_{a2-},t_{a2+}]  &\text{if } x_{a1}\leq  -a^{*}  < a^{*} \leq x_{a2}\\
                            \emptyset&\text{if }   -a^{*}<x_{a1}\leq x_{a2}  < a^{*}   \\
    [t_{a2-},t_{a2+}]  &\text{if }  -a^{*} <x_{a1} < a^{*} \leq x_{a2}\\
    [t_{a2-},t_{a1-}]\cup [t_{a1+},t_{a2+}] & \text{if } a^{*} \leq x_{a1}  \leq x_{a2}
  \end{cases} \label{eq:iahalthb}.
\end{eqnarray}
}
Considering line $b$ we have:
\begin{eqnarray}
  \frac{d x_b}{dt} &<& 0 \forall t\\
  \lim_{t \to 0^-} x_b &=&\lim_{t \to +\infty} x_b =-\infty\\
  \lim_{t \to -\infty} x_b&=&\lim_{t \to 0^+} x_b = +\infty
\end{eqnarray}
and thus
\begin{eqnarray}
  I_{b}=[t_{b2-},t_{b1-}] \cup [t_{b2+},t_{b1+}].
  \label{eq:ibhalthb}
\end{eqnarray}

From the results of this section it follows that the determination of whether an edge of the robot collides with a line during an arc movement can be accomplished without iterative computation. Once $I_{d_i}$ (Eq.\ref{eq:idi}), $I_{a}$ (Eq.\ref{eq:iahagthb}, Eq.\ref{eq:iahaeqhb}, or Eq.\ref{eq:iahalthb}), and $I_{b}$ (Eq.\ref{eq:ibhagthb}, Eq.\ref{eq:ibhaeqhb}, or Eq.\ref{eq:ibhalthb}) are computed, Eq.\ref{eq:intersectionSegment} or Eq.~\ref{eq:intersectionLine} provides a collision test.

\begin{figure}%
\centering
\includegraphics[width=0.9\linewidth]{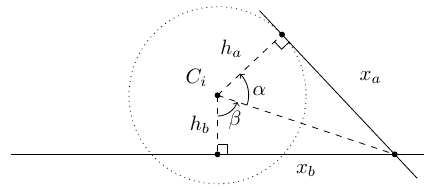}  
\caption{Geometric interpretation of Eq.~\ref{eq:taplus}.}%
\label{fig:taplus}%
\end{figure}

It is worth noting that while the above formulation are more efficient computationally, Eq.~\ref{eq:taplus} to~\ref{eq:tbminus}
can also be interpreted geometrically for a more intuitive understanding. For instance Eq.~\ref{eq:taplus} can be rewritten as:
\begin{eqnarray}
  t_{a+}&=&\frac{x_a + \sqrt{x_a^{2}+h_{a}^{2} - h_{b}^{2}}}{h_{a} + h_{b}}\\
         &=&\frac{\frac{x_a}{\sqrt{x_a^2+h_a^2}} + \frac{\sqrt{x_a^{2}+ h_{a}^{2} - h_{b}^{2}}}{\sqrt{x_a^2+h_a^2}} }{ \frac{h_{a}}{\sqrt{x_a^2+h_a^2}} + \frac{h_{b}}{\sqrt{x_a^2+h_a^2}}}.
\end{eqnarray}
By introducing the angles
\begin{eqnarray}
  \alpha&=&\arcsin\left(\frac{x_a}{\sqrt{x_a^2+h_a^2}}\right)\\
  \beta&=&\arccos\left(\frac{h_{b}}{\sqrt{x_a^2+h_a^2}}    \right)\\
        &=&\arccos\left(\frac{h_{b}}{\sqrt{x_b^2+h_b^2}}    \right)
\end{eqnarray}
and noting that
\begin{eqnarray}
  \tan\left(\frac{\alpha+\beta}{2}   \right) &=&\frac{\sin(\alpha) + \sin(\beta)  }{ \cos(\alpha)+\cos(\beta)  }
\end{eqnarray}
we can interpret $t_{a+}$ as the tangent of half of the angle
\begin{eqnarray}
\theta_{a+} &=& \alpha+\beta.
\end{eqnarray}
This interpretation offers a geometric visualization of how $t_{a+}$ relates to angles and trigonometric functions, as illustrated in Figure~\ref{fig:taplus}.
In a similar manner, $t_{a-}$, $t_{b-}$, and $t_{b+}$ can be interpreted as the tangents of half of the following angles:
{\small\begin{eqnarray}
  \theta_{a-}&=& \phantom{+}\arcsin\left(\frac{x_a}{\sqrt{x_a^2+h_a^2}}\right) - \arccos\left(\frac{h_{b}}{\sqrt{x_a^2+h_a^2}}    \right) \\
  \theta_{b-}&=& -\arcsin\left(\frac{x_b}{\sqrt{x_b^2+h_b^2}}\right) - \arccos\left(\frac{h_{a}}{\sqrt{x_a^2+h_a^2}}    \right)\\
  \theta_{b+}&=& -\arcsin\left(\frac{x_b}{\sqrt{x_b^2+h_b^2}}\right) + \arccos\left(\frac{h_{a}}{\sqrt{x_a^2+h_a^2}}    \right)   
\end{eqnarray}}

By employing a similar transformation, the number of trigonometric operations needed for the computation of $I_{d_i}$ (given in Eq.\ref{eq:tdminus} and Eq.\ref{eq:tdplus}) can be reduced. Specifically, we have:
\begin{eqnarray}
  \tan\left(\frac{-\mu}{2}\right) &=& \frac{-\sin(\mu)}{\cos(\mu)+1}= \frac{-s_{\mu}}{c_{\mu}+1}\\
  \tan\left(\frac{-\mu+\theta_{d_i}}{2}\right) &=& \frac{-s_{\mu}+\sin(\theta_{d_i}) }{c_\mu+\cos(\theta_{d_i}) }.
\end{eqnarray}
where $c_\mu$ and $s_\mu$ are computed as shown in Eq.~\ref{eq:cmu} and Eq.~\ref{eq:smu}, respectively.

\subsubsection{Collision detection for line obstacles, straight movement}
\label{sec:lineSegment}

In line with the method outlined in Section~\ref{sec:pointSegment}, the convex hull of the union of the vertices from both the initial position ($H_{j}$) and the final position ($H'_{j}$) of the hitbox is computed. Let us denote by $K_{i}$, $0 \leq i < N_k$ the vertices of the convex hull (which are a subset of the vertices of $H_{AB}$ defined in Eq~\ref{eq:convexHull}). Let us also order the vertices such that $K_{i}K_{i+1}$ is an edge of the convex hull and define $K_{N_k}=K_{0}$.

If the obstacle is a line segment $S_lS_{l+1}$, the hyperplane separation theorem offers a criterion for collision detection. It is sufficient to find one direction in which the projections of $K_{i}$, $0 \leq i < N_k$  and the projection of $S_lS_{l+1}$ do not overlap. Candidate directions are given by the normals to $K_iK_{i+1}, 0 \leq i \leq N_k$, and the normal to $S_lS_{l+1}$. If intersection occurs for all of these directions, collision is confirmed.
    
If the obstacle is a (infinite) line passing through $S_l$, collision detection simplifies. It is sufficient to check the projection of the points in $K_{i}$ onto the normal of this line, denoted as $\bm{n_{l}}$. Collision occurs when there are points on both side of the line, i.e. when
\begin{eqnarray*}  
  \exists i ,1 \leq i < N_k : \sgn(\bm{n_{l}}^TK_{i} - \bm{n_{l}}^TS_l) \neq \sgn(\bm{n_{l}}^TK_{0} - \bm{n_{l}}^TS_l).
\end{eqnarray*}

\section{Experiments}
\label{sec:experiment}

Among planners leveraging motion primitives, conformal lattice planners~\cite{mcnaughton2011motion} have demonstrated suitability for scenarios involving predetermined routes, where online planning is primarily employed for avoiding unforeseen obstacles. Alternatively, these planners can also be applied to modify a path created by a higher-level planner that operates less frequently. The computation within conformal lattice planners is anticipated to occur onboard in a reactive manner. This scenario is expected to greatly benefit from motion primitives requiring low computational effort. Therefore, we selected conformal lattice planners as the testbed for our motion primitives.

Specifically, let us assume the predefined path comprises a series of $N_W$ waypoints, with each waypoint being defined by a 2D position $\bm{w}_{i,0} \in \mathbb{R}^2$ and an orientation $\theta_{i} \in \mathbb{R}$, $i \in \mathbb{N}$, $1 \leq i \leq N_W$.
To create a lattice from this initial set of poses, we establish positions $\bm{w}_{i,j}$ according to the formula:
\begin{eqnarray}  
\bm{w}_{i,j}= \bm{w}_{i,0}+ j s \left[  \begin{matrix} -\sin(\theta_i) \\ \phantom{-}\cos(\theta_i) \end{matrix} \right]
\end{eqnarray}
where $j \in \mathbb{Z}$ and  $s \in \mathbb{R}$  represents the step size (in our implementation, $s=0.2m$). Additionally, we assume that all positions $\bm{w}_{i,j}$ should be reached with the same orientation $\theta{i}$.

In the lattice, each node $(i,j)$, where $1 \leq i \leq N_W$, corresponds to a physical world waypoint denoted as $\bm{w}_{i,j}$. Here, we assume that $-h \leq j \leq h$, with $h$ representing the maximum expansion of the lattice on each side. We permit only edges in the form of $(i,j)$ to $(i+1,k)$ to exist. To these edges, we assign a cost of $1+|j|+|k|$ if the edge is collision-free, and infinite cost if it leads to collisions. This cost assignment gives preference to collision-free paths that closely follow the provided waypoints.

Classic A* is used to perform planning on this lattice. For each node $(i,j)$ in the lattice, the heuristic function is set as $N_W-i +|j|$. We assume the current robot position is connected  to all nodes $(i_b,j)$, with $-h \leq j \leq h$. The value of $i_b \in \mathbb{N}$, with $1 \leq i_b \leq N_W$, is updated based on the robot's progress along the lattice. All nodes $(i_e,j)$, where $i_e \in \mathbb{N}$, $i_b \leq i_e \leq N_W$, and $-h \leq j \leq h$, are designated as goal nodes. The specific value of $i_e$ depends on the planning horizon, which is determined as follows: If the distance from $\bm{w}{i_b,0}$ to $\bm{w}{N_W,0}$ is less than $d_h$, then $i_e=N_W$. Otherwise, $i_e$ is set as the minimum $i$ for which the distance from $\bm{w}{i_b,0}$ to $\bm{w}{i,0}$ is greater than $d_h$, where $d_h \in \mathbb{R}$ denotes the planning horizon.

The search is carried out iteratively with increasing values of $h$ until a solution is discovered or a predefined threshold is reached. In our implementation, this threshold is set to 15. Since each graph is a supergraph of the previous one, caching of the edge costs is implemented to expedite computation.

Additional speed improvements are achieved by assuming that the environment experiences limited changes between consecutive planning requests. This means that the previous planning is assumed to be still valid unless there are alterations in the waypoints, which might be dynamic and provided by an additional global planner.

In detail, we update the values of $i_b$ and $i_e$ to $i_b'$ and $i_e'$ based on the most recent robot position. Subsequently, a new path is composed by concatenating the following components:
\begin{itemize}
\item the biarc from the current position to the previously selected $\bm{w}_{i_b',j_b'}$, computed using the algorithm explained in Section~\ref{sec:replanJ}, using the arc leading to $\bm{w}_{i_b',j_b'}$ as a reference.
\item  all the biarcs of the previous plan from  $\bm{w}_{i_b',j_b'}$ to $\bm{w}_{i_e,j_e}$
\item (possibly) new biarcs from that connect  $\bm{w}_{i_e,j_e}$ to $\bm{w}_{i_e',0}$ passing through  $\bm{w}_{i,0}$ for ${i_e<i<i_e'}$.
\end{itemize}

The revised trajectory is subject to collision checking, and if no collisions are detected, it is accepted as the new plan. In case of collisions, a fresh plan is generated using the A* method as previously outlined. This strategy has the disadvantage of potentially yielding suboptimal paths when dealing with obstacles that transiently intersect the robot's path. Nevertheless, it offers significant computational speed enhancements and helps mitigate the abrupt and erratic changes in movement that arise from frequent switching between multiple paths due to minor variations in obstacle positions as detected by the LIDAR sensors.

To ensure easily comparable and reproducible outcomes, the planner described previously was assessed using MRPB, a publicly available benchmark for local mobile robot planning detailed in~\cite{wen2021mrpb}. The evaluation employed default settings for the test robot, DWA and TEB parameters, and other relevant configurations available in the standard MRPB testing suite. The robot's hitbox was configured as a square with dimensions of 34 cm within our planner. Waypoints were generated by sampling the path provided by the global planner at 0.5 m intervals. The planning horizon was set to 25 m for our planner. All experiments were conducted on a workstation featuring an Intel I9-9900K processor operating at 3.6 GHz, coupled with 32 GB of RAM.

The results are presented in Table~\ref{tab:mrpbResults}. The columns provide information on the environment (map), the initial and goal positions (test), the utilized planner, the planning time (denoted as $C$ in~\cite{wen2021mrpb}), the path length (represented as $S$ in~\cite{wen2021mrpb}), the time required to reach the goal (referred to as $T$ in~\cite{wen2021mrpb}), and the closest distance of the robot to the obstacles (listed under the \emph{prox} column and corresponding to $d_o$ in~\cite{wen2021mrpb}).

The results indicate that the planning time of our approach is significantly faster by two orders of magnitude compared to commonly used ROS planners. Although the path length in our planner is generally longer, the time it takes to reach the goal is comparable. It is worth emphasizing that many ROS planners optimize path length by taking shortcuts with respect to the global plan. This shortcut behavior is undesirable in many specific applications, where maintaining proximity to the global plan is crucial.

\begin{table}
  \caption{Evaluation Results for MRPB 1.0}
  \label{tab:mrpbResults}
  \resizebox{\columnwidth}{!}{%
    \begin{tabular}{lrlrrrr}
\hline
\toprule
map & test & planner & plan [ms] & path [m] & time [s] & prox [m] \\
\midrule
\hline
maze & 1 & biarc & 0.013 & 46.04 & 119.84 & 0.32 \\
maze & 1 & dwa & 7.995 & 40.67 & 92.87 & 0.28 \\
maze & 1 & teb & 6.473 & 43.06 & 114.28 & 0.33 \\
\hline
maze & 2 & biarc & 0.012 & 43.82 & 94.78 & 0.40 \\
maze & 2 & dwa & 7.588 & 40.34 & 89.92 & 0.27 \\
maze & 2 & teb & 6.893 & 42.15 & 107.94 & 0.31 \\
\hline
maze & 3 & biarc & 0.019 & 46.21 & 103.76 & 0.23 \\
maze & 3 & dwa & 7.533 & 40.15 & 96.19 & 0.29 \\
maze & 3 & teb & 6.927 & 42.53 & 117.83 & 0.33 \\
\hline
narrow graph & 1 & biarc & 0.015 & 32.10 & 68.60 & 0.33 \\
narrow graph & 1 & dwa & 6.737 & 28.76 & 67.29 & 0.27 \\
narrow graph & 1 & teb & 11.920 & 30.99 & 87.92 & 0.33 \\
\hline
narrow graph & 2 & biarc & 0.015 & 32.14 & 68.97 & 0.29 \\
narrow graph & 2 & dwa & 7.538 & 28.24 & 62.00 & 0.26 \\
narrow graph & 2 & teb & 8.212 & 29.44 & 71.87 & 0.32 \\
\hline
narrow graph & 3 & biarc & 0.012 & 29.07 & 68.67 & 0.26 \\
narrow graph & 3 & dwa & 6.562 & 25.29 & 62.05 & 0.28 \\
narrow graph & 3 & teb & 12.503 & 26.66 & 73.04 & 0.37 \\
\hline
office01add & 1 & biarc & 0.005 & 18.54 & 41.35 & 0.39 \\
office01add & 1 & dwa & 8.017 & 17.78 & 40.87 & 0.30 \\
office01add & 1 & teb & 7.185 & 18.23 & 42.14 & 0.38 \\
\hline
office01add & 2 & biarc & 0.778 & 17.87 & 49.63 & 0.28 \\
office01add & 2 & dwa & 6.872 & 16.00 & 42.90 & 0.32 \\
office01add & 2 & teb & 7.236 & 16.88 & 45.29 & 0.31 \\
\hline
office01add & 3 & biarc & 0.244 & 16.10 & 39.66 & 0.25 \\
office01add & 3 & dwa & 7.161 & 15.32 & 43.51 & 0.26 \\
office01add & 3 & teb & 10.228 & 16.80 & 46.08 & 0.31 \\
\hline
office02 & 1 & biarc & 0.003 & 30.00 & 74.46 & 0.39 \\
office02 & 1 & dwa & 9.075 & 29.05 & 71.67 & 0.28 \\
office02 & 1 & teb & 4.625 & 29.49 & 73.99 & 0.30 \\
\hline
office02 & 2 & biarc & 0.006 & 32.77 & 77.74 & 0.33 \\
office02 & 2 & dwa & 8.223 & 31.65 & 79.76 & 0.27 \\
office02 & 2 & teb & 5.845 & 32.39 & 86.62 & 0.31 \\
\hline
office02 & 3 & biarc & 0.006 & 35.85 & 88.43 & 0.34 \\
office02 & 3 & dwa & 8.975 & 34.26 & 83.00 & 0.31 \\
office02 & 3 & teb & 4.074 & 35.24 & 85.51 & 0.39 \\
\hline
room02 & 1 & biarc & 1.005 & 17.93 & 41.68 & 0.28 \\
room02 & 1 & dwa & 8.102 & 15.99 & 36.96 & 0.34 \\
room02 & 1 & teb & 4.916 & 16.47 & 37.54 & 0.33 \\
\hline
room02 & 2 & biarc & 0.029 & 16.10 & 37.83 & 0.35 \\
room02 & 2 & dwa & 7.274 & 14.02 & 34.68 & 0.29 \\
room02 & 2 & teb & 7.382 & 15.13 & 45.15 & 0.34 \\
\hline
room02 & 3 & biarc & 0.013 & 14.42 & 32.56 & 0.44 \\
room02 & 3 & dwa & 9.383 & 13.28 & 31.52 & 0.30 \\
room02 & 3 & teb & 4.559 & 13.58 & 29.96 & 0.33 \\
\hline
shopping mall & 1 & biarc & 0.010 & 50.22 & 124.37 & 0.46 \\
shopping mall & 1 & dwa & 9.275 & 46.66 & 113.10 & 0.30 \\
shopping mall & 1 & teb & 4.394 & 48.27 & 123.12 & 0.37 \\
\hline
shopping mall & 2 & biarc & 0.021 & 53.39 & 134.01 & 0.39 \\
shopping mall & 2 & dwa & 8.564 & 49.44 & 122.02 & 0.31 \\
shopping mall & 2 & teb & 4.279 & 50.88 & 127.17 & 0.35 \\
\hline
shopping mall & 3 & biarc & 0.004 & 50.13 & 122.30 & 0.50 \\
shopping mall & 3 & dwa & 8.740 & 48.33 & 118.52 & 0.33 \\
shopping mall & 3 & teb & 3.964 & 49.39 & 118.38 & 0.37 \\
\hline
six people & 1 & biarc & 0.048 & 13.94 & 34.90 & 0.33 \\
six people & 1 & dwa & 7.106 & 13.20 & 34.00 & 0.28 \\
six people & 1 & teb & 7.825 & 14.17 & 37.85 & 0.25 \\
\hline
track & 1 & biarc & 0.004 & 74.83 & 161.72 & 0.21 \\
track & 1 & dwa & 8.000 & 69.84 & 146.38 & 0.28 \\
track & 1 & teb & 9.471 & 72.42 & 165.67 & 0.32 \\
\hline
\bottomrule
\end{tabular}

  }
\end{table}

\section{Conclusions}
\label{sec:conclusions}

In this paper, our primary focus was on enhancing planning efficiency by strategically choosing motion primitives. More specifically, we opted for biarcs as our motion primitives, enabling collision detection without the need for extensive sampling of the robot's trajectory.

We conducted a review of biarcs, paying particular attention to how the selection of the joint point location impacts both path length and curvature discontinuity. Our investigation led us to identify the equal-chord biarc as a favorable compromise between these two factors. Additionally, we introduced a heuristic for determining the joint location when we need a biarc to approximate another pre-defined biarc, which is a common scenario in the incremental updating of plans.

Subsequently, we presented collision detection criteria for robot paths composed of biarcs, considering scenarios with a polygonal hitbox and three types of obstacles: points, line segments, and lines. Our analysis demonstrated that each case demands very limited computational resources.

We conducted an analysis of a conformal lattice motion planner paired with biarc motion primitives within a publicly available benchmarking suite for local motion planners. Our findings indicate that our planner offers performance comparable to commonly used open-source motion planners while significantly reducing computation time.

A significant limitation of the current approach is its exclusive focus on collision detection without considering the minimum distance to obstacles. Depending on the application, incorporating this information in the weights of graph edges may be required. Future work should prioritize extending the results presented in Section~\ref{sec:collisionDetection} to compute this additional information.

Furthermore, our current approach only considers static objects, and while it is a reasonable assumption in our application, it is crucial to explore this extensions in other fields, such as self-driving cars that navigate in public road traffic.

Another limitation to address is that biarcs are only $G_1$ continuous. This discontinuity in curvature at the joint point could be problematic for Ackermann steering platforms and fast-moving robots, potentially imposing undesired stress on the powertrain. One promising improvement is to replace the central section of the biarc with a clothoid. By limiting the length of the clothoid section, it may be possible to reduce the number of robot positions that need to be sampled and checked for collisions. Future work will involve testing the performance of this solution, which is expected to provide smoother curves without significantly increasing planning time.

\ifCLASSOPTIONcaptionsoff
  \newpage
\fi

\bibliographystyle{IEEEtran}
\bibliography{IEEEabrv,collisionAvoidance,biarc}

\end{document}